\PassOptionsToPackage{table}{xcolor}
\documentclass{fairmeta}

\usepackage{soul}
\usepackage{titletoc}
\usepackage{listings}
\usepackage{float}
\usepackage{tabularx}
\usepackage{array}
\usepackage{colortbl}
\usepackage{makecell}

\definecolor{riseblue}{RGB}{47,92,153}
\definecolor{risebluebg}{RGB}{232,240,250}

\definecolor{risegreen}{RGB}{48,132,88}
\definecolor{risegreenbg}{RGB}{226,243,233}

\definecolor{risebluehl}{RGB}{207,232,255}

\definecolor{risegold}{RGB}{178,126,24}
\definecolor{risegoldbg}{RGB}{255,244,204}

\definecolor{risered}{RGB}{174,63,63}
\definecolor{riseredbg}{RGB}{252,229,229}

\definecolor{risegray}{RGB}{95,101,110}
\definecolor{risegraybg}{RGB}{244,245,247}

\definecolor{oursbg}{RGB}{226,242,228}

\newcommand{\hlshared}[1]{%
  {\sethlcolor{risegoldbg}\hl{#1}}%
}

\newcommand{\hlextra}[1]{%
  {\sethlcolor{risebluehl}\hl{#1}}%
}

\newcommand{\hlstrength}[1]{%
  {\sethlcolor{risegreenbg}\hl{#1}}%
}

\newcommand{\hlfailure}[1]{%
  {\sethlcolor{riseredbg}\hl{#1}}%
}

\newtcolorbox{promptbox}[1][]{%
  breakable,
  enhanced,
  colback=risebluebg,
  colframe=riseblue,
  coltitle=white,
  fonttitle=\bfseries,
  boxrule=0.7pt,
  arc=1.2mm,
  left=6pt,
  right=6pt,
  top=5pt,
  bottom=5pt,
  attach boxed title to top left={
    xshift=5mm,
    yshift=-2mm
  },
  boxed title style={
    colback=riseblue,
    colframe=riseblue,
    boxrule=0pt,
    arc=1mm
  },
  #1
}

\DeclareCaptionStyle{ruled}{
  labelfont=normalfont,
  labelsep=colon,
  strut=off
}

\floatstyle{ruled}
\newfloat{listing}{tb}{lst}{}
\floatname{listing}{Listing}

\newcommand{\resultcell}[2]{%
  \makebox[2.55em][r]{#1}%
  \makebox[2.20em][l]{#2}%
}

\newcommand{\initres}[1]{%
  \resultcell{#1}{}%
}

\newcommand{\gain}[2]{%
  \resultcell{#1}{%
    \raisebox{-0.45ex}{%
      \scriptsize
      \textcolor{green!55!black}{$+\!#2$}%
    }%
  }%
}

\newcommand{\loss}[2]{%
  \resultcell{#1}{%
    \raisebox{-0.45ex}{%
      \scriptsize
      \textcolor{red!75!black}{$-\!#2$}%
    }%
  }%
}

\newcommand{\res}[2]{\( #1 \pm #2 \)}
\newcommand{\best}[2]{\(\mathbf{#1 \pm #2}\)}

\usepackage{titletoc}

\usepackage{algorithm}
\usepackage{algpseudocode}

\usepackage{amsmath,amsfonts,bm}

\def\eqref#1{equation~\ref{#1}}

\def\1{\bm{1}}

\DeclareMathAlphabet{\mathsfit}{\encodingdefault}{\sfdefault}{m}{sl}
\SetMathAlphabet{\mathsfit}{bold}{\encodingdefault}{\sfdefault}{bx}{n}

\usepackage{url}
\usepackage{amsmath,amssymb}
\usepackage{wrapfig}
\usepackage{tabularx}
\usepackage{makecell}
\usepackage{colortbl}
\usepackage{pifont}

\setcitestyle{round,authoryear}

\usepackage{adjustbox}

\headerleft{\textbf{RISE-RL}}
\headerright{}

\title{\Large RISE-RL: Rubric-Informed Selective Exploration for Open-Ended Reinforcement Learning}

\author[1,\ddagger]{Jinkun Hou}
\author[2,\ddagger]{Zhuo Liu}
\author[3,\dagger]{Huimin Ren}
\author[4]{Kuan Zhang}
\author[4]{Zirong Chen}
\author[3]{Hongsheng Xin}
\author[3]{Pan Zhou}
\author[3]{Kun Zhan}
\renewcommand\affiliationformat[2][]{{\small $^{#1}$#2}}
\affiliation[1]{Peking University}
\affiliation[2]{Beijing Institute of Technology}
\affiliation[3]{Li Auto Inc.}
\affiliation[4]{Tsinghua University}
\contribution[\dagger]{Corresponding Author}
\contribution[\ddagger]{Work done while interning at Li Auto Inc.}

\abstract{Aligning Large Language Models (LLMs) for open-ended tasks is challenging because responses must satisfy multidimensional criteria without following a single correct generation trajectory. Existing rubric-based reinforcement learning (RL) methods compress fine-grained criterion-level feedback into scalar rewards, making persistent capability gaps difficult to target under limited on-policy exploration. We propose \textbf{RISE-RL} (Rubric-Informed Selective Exploration), which uses frequently missed rubric criteria under natural rollouts to elicit privileged trajectories that are difficult to discover through unguided exploration alone. RISE-RL retains only trajectories whose complete-rubric reward exceeds the mean reward of natural rollouts, and then re-evaluates them under the original prompt to emphasize behaviors that remain weakly supported by the natural policy. The resulting guidance signal is optimized through a separate auxiliary objective and removed once its additional benefit diminishes. Experiments with 4B and 14B models across writing, chat, health, and science show that RISE-RL achieves the highest mean score on every evaluated benchmark under guidance-free evaluation. Compared with standard Rubric-RL, it improves the average score by 1.3 points at the 4B scale and \textbf{3.3 points at the 14B scale}, including a \textbf{6.0-point} gain on CreativeWriting-V3. It also improves creative-writing diversity and yields gains on objectively scored medical and scientific benchmarks. These results indicate that selective internalization through reward filtering and policy support shaping is effective for open-ended reinforcement learning.

}

\metadata[Email]{\email{houjinkun26@stu.pku.edu.cn}, \email{renhuimin@lixiang.com}}

\begin{document}

\maketitle

\section{Introduction}

\begin{figure}[t]
\centering
\includegraphics[width=\columnwidth]{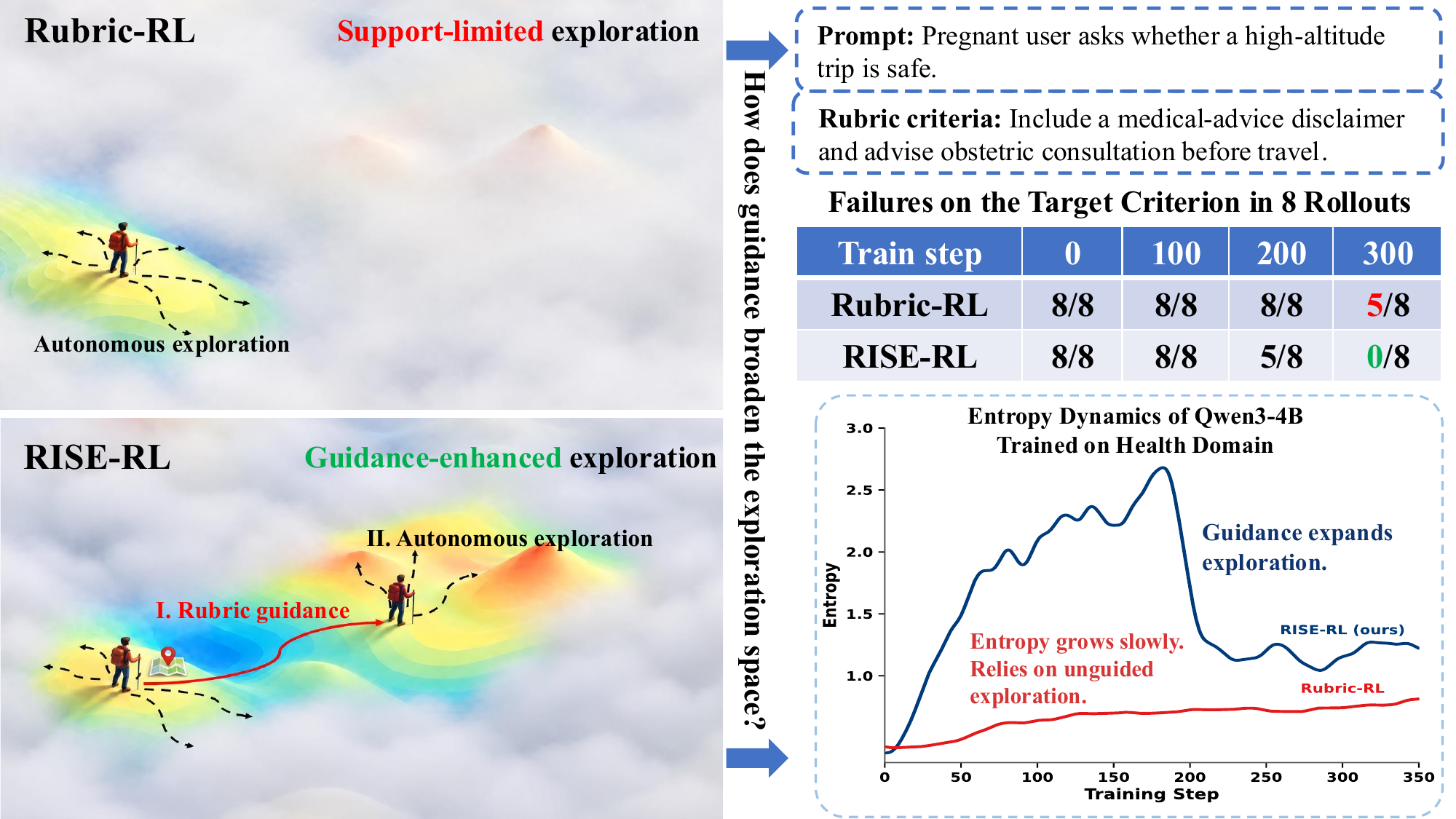} 

\caption{Overview of the guidance--exploration mechanism in RISE-RL. Selective rubric guidance addresses criterion failures and expands the accessible policy space. Once these behaviors become accessible, guidance is removed and training continues with autonomous on-policy exploration.}
\label{fig1}
\end{figure}

Open-ended question answering is a core capability of large language models, spanning open-domain dialogue~\citep{zhang2020dialogpt,roller2021recipes,thoppilan2022lamda}, creative writing~\citep{franceschelli2023creativity,gomez2023confederacy}, health assistance~\citep{lin2025healthgpt,singhal2023publisher,singhal2025toward}, and scientific explanation~\citep{wan2024sciqag,besrour2025squai}. Unlike mathematical reasoning~\citep{ren2025deepseek,chen2025seed} or code generation~\citep{le2022coderl,cao2026qwen3}, open-ended tasks admit no single correct trajectory and require responses to satisfy diverse criteria, including factuality, completeness, style, structure, safety, and human preferences. Moreover, substantially different responses may be equally valid, giving rise to a multi-peaked policy distribution with multiple desirable generation modes. Reinforcement learning for open-ended tasks therefore faces the challenge of improving rubric alignment without sacrificing exploration or response diversity. Balancing these objectives remains a central challenge.

Existing methods still struggle to improve open-ended capabilities while preserving exploration and diversity. Rubric-based RL provides fine-grained criterion-level feedback but aggregates criterion-wise scores into a scalar sequence-level reward for policy optimization~\citep{gunjal2025rubrics,huang2026sam}. Consequently, criterion-specific failures may not be directly translated into targeted policy updates. OPD provides dense token-level supervision by distilling a teacher distribution on states visited by the student policy~\citep{li2026rethinking,zhao2026rosd,hubotter2026reinforcement}. While such supervision facilitates knowledge transfer, it offers limited control over which guided behaviors are selectively internalized. Mixed-policy optimization further combines privileged trajectories with on-policy samples in the same group-relative objective~\citep{bi2025reward,huang2026think}. This coupling alters the relative advantages of natural rollouts and can cause scaffold-removal collapse, a performance drop when generation returns to natural prompts.

To address these limitations, we propose RISE-RL, a selective-guidance paradigm for open-ended tasks. RISE-RL uses rubric feedback to identify capability gaps, filters guided trajectories by complete-rubric reward, and estimates their natural-policy support---the token probabilities assigned under the original prompt---to emphasize beneficial behaviors that remain unlikely without guidance. The resulting signal is optimized through a separate auxiliary objective, keeping guided learning outside natural group-relative optimization. As its benefit diminishes, guidance is removed and training continues with unguided on-policy exploration, facilitating selective internalization while retaining autonomous exploration and response diversity.

We evaluate RISE-RL on Qwen3-4B and Qwen3-14B across four RubricHub domains and eight downstream benchmarks. RISE-RL achieves higher mean scores than standard Rubric-RL on every evaluated benchmark at both model scales, with average gains of \textbf{1.3} and \textbf{3.3} points at the 4B and 14B scales, respectively. On Qwen3-14B, the gains reach \textbf{8.0 points} on Arena-Hard-v2 and \textbf{6.0 points} on CreativeWriting-V3. In the creative-writing evaluation, RISE-RL also improves output diversity by \textbf{5.8\%}. These gains further extend to objectively scored tasks, including improvements of \textbf{3.3 points} on MedQA and \textbf{3.6 points} on GPQA-Diamond. Overall, these results suggest that RISE-RL improves complex rubric alignment while supporting generation diversity and facilitating the internalization of domain knowledge and reasoning capabilities.

Our main contributions, which directly address the limitations of
prior work, are threefold:

\begin{itemize}
    \item \textbf{Criterion-Level Selective Guidance for Open-Ended RL:} We introduce RISE-RL, which retains the fine-grained diagnostic information that is normally collapsed into a scalar rubric reward. By identifying high-value criteria that remain repeatedly unsatisfied across natural rollouts, RISE-RL constructs targeted guidance for behaviors that are rarely elicited through unguided exploration.

    \item \textbf{Selective Internalization by Gain and Policy Support:} RISE-RL retains only privileged trajectories that improve the complete-rubric reward over the natural-rollout baseline. It then removes the privileged criteria and re-evaluates the retained trajectories under the original prompt, using token probabilities to concentrate learning on beneficial behaviors that remain weakly supported by the natural policy. The resulting guidance signal is optimized through a separate auxiliary objective, preventing privileged trajectories from affecting the reward normalization and relative-advantage estimation of natural on-policy rollouts.

    \item \textbf{Broad Empirical Evaluation across Models and Domains:} Experiments with Qwen3-4B and Qwen3-14B across writing, chat, health, and science show that RISE-RL achieves higher mean scores than Rubric-RL on every evaluated benchmark at both scales under guidance-free evaluation. The gains are particularly pronounced on challenging open-ended benchmarks at the 14B scale, while improvements on objectively scored tasks further suggest within-domain transfer to verifiable tasks. Training-dynamics analyses and ablations additionally support the benefits of selective early guidance followed by unguided on-policy exploration.
\end{itemize}
\section{Related Work}

\paragraph{Rubric-Based RL and OPD.}
Rubric-based RL~\citep{gunjal2025rubrics} characterizes the multidimensional quality of open-ended responses through fine-grained evaluation criteria and aggregates criterion-level scores into a scalar reward for policy optimization. However, this aggregation can discard part of the diagnostic information provided by the judge, making failures on specific criteria difficult to translate into targeted policy updates. A complementary line of work instead provides dense supervision through policy distillation. On-policy distillation (OPD) queries a teacher distribution on states visited by the student policy~\citep{li2026rethinking}, while on-policy self-distillation (OPSD) constructs the teacher using the same model under privileged contexts~\citep{zhao2026self}. Rubric-Guided Self-Distillation further supplies instance-specific rubrics as privileged information to the teacher~\citep{rezaei2026rubric}. Although these methods offer token-level guidance, their objectives primarily emphasize matching the teacher distribution, without explicitly coordinating dense supervision with the preservation of autonomous policy exploration.

\paragraph{Guided Trajectory Learning for Policy Optimization.}
Recent work has begun to combine natural on-policy rollouts with externally guided trajectories. LUFFY integrates on-policy rollouts with off-policy expert demonstrations and introduces policy shaping to balance imitation and exploration~\citep{yan2026learning}. Critique-GRPO~\citep{zhang2025critique} and RGR-GRPO~\citep{bi2025reward} further construct refined responses from critiques and rubrics, respectively, and jointly optimize them with the original rollouts. RuscaRL instead injects rubric-based scaffolds into rollout prompts, producing trajectories conditioned on additional guidance~\citep{zhou2025breaking}. These methods demonstrate that guided trajectories can expose the policy to high-quality behaviors that are difficult to discover through autonomous exploration alone. However, guided and natural trajectories often share the same optimization objective, coupling the learning signal from privileged trajectories with the relative advantages of natural samples. This coupling may reduce control over their respective contributions and exacerbate distribution shift when the scaffold is removed. RISE-RL differs by filtering guided trajectories with the complete rubric, estimating their policy support after removing the privileged context, and selectively internalizing them through a separate auxiliary objective.

\section{Method}

\begin{figure*}[t]
\centering
\includegraphics[width=\textwidth]{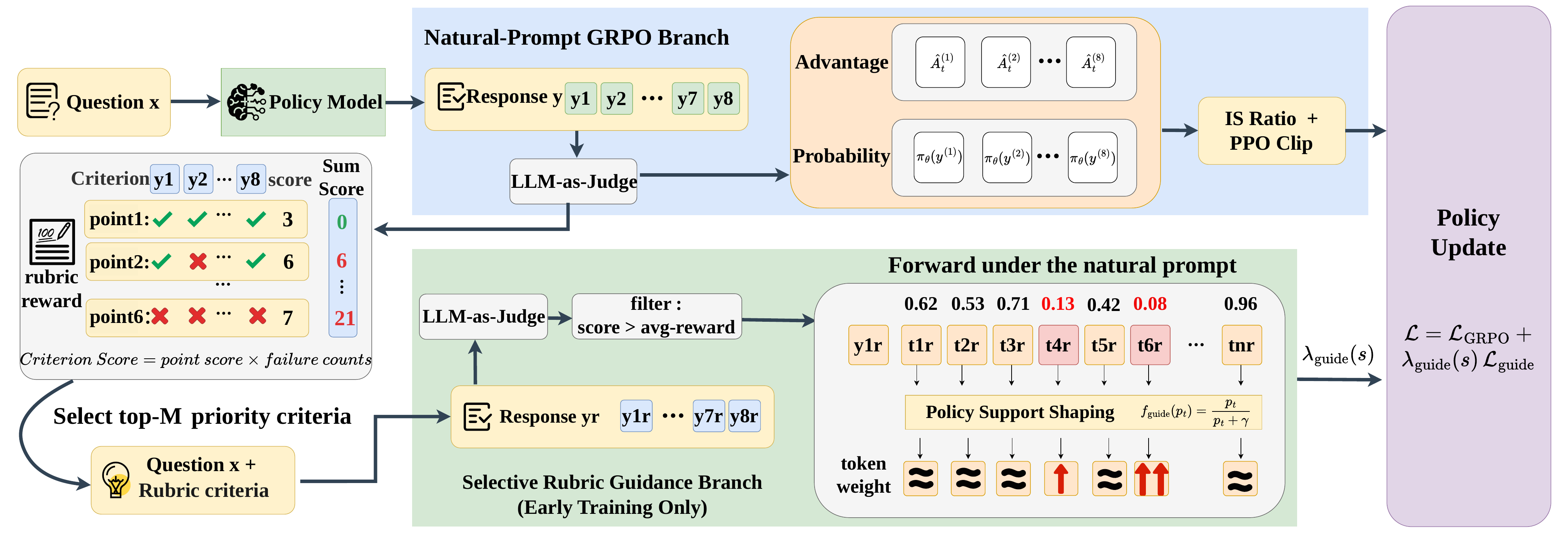} 

\caption{Overview of RISE-RL. Alongside natural-prompt GRPO, an early selective-guidance branch targets unmet rubric criteria and reinforces low-support tokens in reward-improving trajectories. Guidance is removed once its reward advantage saturates, after which training continues with pure GRPO.}
\label{fig2}
\end{figure*}

\subsection{Group Relative Policy Optimization}

We build on Group Relative Policy Optimization (GRPO)~\citep{shao2024deepseekmath} for policy optimization under the unguided prompt. We refer to this as the natural-rollout branch. Given a prompt $q$, the old policy $\pi_{\theta_{\mathrm{old}}}$ samples a group of $G$ responses $\{o_i\}_{i=1}^{G}$, each assigned a sequence-level reward $r_i$. GRPO estimates the advantage of each response by normalizing its reward within the group:
\begin{equation}
\hat{A}_i
=
\frac{
r_i-\operatorname{mean}\!\left(\{r_j\}_{j=1}^{G}\right)
}{
\operatorname{std}\!\left(\{r_j\}_{j=1}^{G}\right)+\epsilon_A
},
\label{eq:grpo_advantage}
\end{equation}
where $\epsilon_A$ is a small constant for numerical stability. The resulting sequence-level advantage is shared by all tokens in $o_i$. We instantiate the natural rollout branch with GRPO, following standard implementation practice.

\subsection{Criterion-Level Selective Feedback}

For each prompt $q$, we associate a rubric $\mathcal{C}(q)=\{c_k\}_{k=1}^{K}$, where criterion $c_k$ has weight $w_k$. For response $o_i$, the rubric grader
produces a binary judgment
\begin{equation}
z_{i,k}
=
\mathbb{I}\!\left[o_i \text{ satisfies } c_k\right],
\qquad
z_{i,k}\in\{0,1\},
\label{eq:criterion_judgment}
\end{equation}
and GRPO uses the aggregated sequence-level reward
\begin{equation}
r_i
=
\frac{\sum_{k=1}^{K}w_k z_{i,k}}
{\sum_{k=1}^{K}w_k}.
\label{eq:rubric_reward}
\end{equation}

Although suitable for policy optimization, this scalar reward obscures which important criteria are repeatedly missed across natural rollouts. Given
$\mathcal{O}^{\mathrm{nat}}
=\{o_i^{\mathrm{nat}}\}_{i=1}^{G}$,
we therefore define the priority of criterion $c_k$ as
\begin{equation}
p_k
=
w_k
\sum_{i=1}^{G}
\left(
1-z_{i,k}^{\mathrm{nat}}
\right),
\label{eq:criterion_priority}
\end{equation}
which jointly captures criterion importance and failure frequency within the current rollout group. We then select the $M$ highest-priority criteria as targeted feedback:
\begin{equation}
\mathcal{C}^{\mathrm{fb}}(q)
=
\operatorname{TopM}
\left(
\mathcal{C}(q);
\{p_k\}_{k=1}^{K}
\right).
\label{eq:targeted_feedback}
\end{equation}

We select at most $M$ criteria with positive priority; if fewer than $M$ have $p_k>0$, we use only those criteria, and skip guidance when none exists. This concentrates guidance on high-value capabilities that remain insufficiently covered by the current policy.

\subsection{Quality-Filtered Selective Guidance}

We append the selected criteria to the original prompt and sample a group of privileged trajectories:
\begin{equation}
\begin{aligned}
q^{\mathrm{priv}}
&=
q\oplus\mathcal{C}^{\mathrm{fb}}(q),
\\
o_j^{\mathrm{priv}}
&\sim
\pi_{\theta_{\mathrm{old}}}
\left(
\cdot\mid q^{\mathrm{priv}}
\right),
\qquad j=1,\ldots,G.
\end{aligned}
\label{eq:privileged_rollouts}
\end{equation}

Because privileged conditioning may cause cross-criterion degradation or prompt mismatch, we re-evaluate each candidate using the complete rubric; this filtering reduces, but does not eliminate, these risks. Let $\bar{r}^{\mathrm{nat}}$ and
$\sigma^{\mathrm{nat}}$ denote the mean and standard deviation of the
natural-rollout rewards $\{r_i^{\mathrm{nat}}\}_{i=1}^{G}$, respectively.
We define the normalized reward gain of each privileged trajectory as
\begin{equation}
A_j^{\mathrm{ref}}
=
\frac{r_j^{\mathrm{priv}}-\bar{r}^{\mathrm{nat}}}
{\sigma^{\mathrm{nat}}+\epsilon_A},
\qquad \epsilon_A=10^{-6}.
\end{equation}

To identify important yet under-supported tokens in reward-improving trajectories, we remove the selected criteria and teacher-force each reference trajectory under the original prompt:
\begin{equation}
p_{j,t}
=
\pi_\theta
\left(
o_{j,t}^{\mathrm{ref}}
\mid
q,o_{j,<t}^{\mathrm{ref}}
\right).
\label{eq:natural_token_probability}
\end{equation}
Re-evaluating the retained trajectory under the original prompt estimates its support without privileged criteria: a low $p_{j,t}$ indicates that the corresponding token, despite appearing in a higher-reward trajectory, remains weakly supported by the natural policy. Following the saturating transformation in LUFFY~\citep{yan2026learning}, we define the policy support factor
\begin{equation}
\rho_{j,t}
=
\frac{p_{j,t}}{p_{j,t}+\gamma},
\label{eq:familiarity_factor}
\end{equation}
where $\gamma>0$ controls the sensitivity to low-probability tokens. As $p_{j,t}$ decreases, $-\log\rho_{j,t}$ increases, assigning stronger guidance to weakly supported behavior while suppressing updates to already-supported tokens.

For each retained trajectory
$o_j^{\mathrm{ref}}\in\mathcal O^{\mathrm{ref}}$, let $T_j$
denote its token length. We define the trajectory-level weight as
$\widetilde A_j=\min\!\left(A_j^{\mathrm{ref}},A_{\max}\right)$,
where $A_{\max}=5$ limits large normalized gains when the
natural-rollout reward variance is small.
The selective guidance loss is then
\begin{equation}
\mathcal{L}_{\mathrm{guide}}
=
-
\frac{1}{|\mathcal O^{\mathrm{ref}}|}
\sum_{o_j^{\mathrm{ref}}\in\mathcal O^{\mathrm{ref}}}
\frac{\widetilde A_j}{T_j}
\sum_{t=1}^{T_j}
\log \rho_{j,t}.
\label{eq:selective_guidance_loss}
\end{equation}

This objective combines two forms of selectivity: reward filtering removes non-improving privileged trajectories, while policy support weighting focuses learning on higher-reward behaviors weakly supported by the current policy. Together, reward-based trajectory selection and probability-adaptive token weighting enable selective internalization of guided responses. We recompute $p_{j,t}$ under the current policy $\pi_\theta$ and backpropagate only through these token probabilities, treating trajectories, rewards, and trajectory weights as fixed. If $\mathcal{O}^{\mathrm{ref}}=\emptyset$, we set $\mathcal{L}_{\mathrm{guide}}=0$.

\subsection{Decoupled Optimization and Guidance Removal}

We incorporate the filtered guidance signal through a separate auxiliary objective:
\begin{equation}
\mathcal{L}(\theta,s)
=
\mathcal{L}_{\mathrm{GRPO}}(\theta)
+
\lambda_{\mathrm{guide}}(s)\mathcal{L}_{\mathrm{guide}}(\theta),
\label{eq:decoupled_objective}
\end{equation}
where $s$ is the training step. Natural rollouts are optimized by GRPO, while filtered privileged trajectories contribute only through the auxiliary loss. This keeps them outside the reward normalization and relative-advantage estimation of natural rollouts, while allowing independent control over guidance strength and token-level weighting.

As the policy internalizes the guided behaviors, the reward advantage of privileged rollouts over natural rollouts gradually decreases. We measure this additional benefit by
\begin{equation}
\begin{aligned}
\Delta_r(s)
&=
\bar r^{\mathrm{priv}}(s)
-
\bar r^{\mathrm{nat}}(s),\\
\bar r^{\mathrm{priv}}(s)
&=
\frac{1}{G}\sum_{j=1}^{G}r_j^{\mathrm{priv}}(s),
\qquad
\bar r^{\mathrm{nat}}(s)
=
\frac{1}{G}\sum_{i=1}^{G}r_i^{\mathrm{nat}}(s).
\end{aligned}
\label{eq:reward_gap}
\end{equation}
A large $\Delta r(s)$ indicates that privileged feedback still elicits behaviors less accessible to the natural policy, whereas a narrowing gap suggests diminishing marginal benefit.

Based on a preliminary run, we set $s_{\mathrm{switch}}$ near the onset of the plateau in the smoothed reward-gap curve and use $\lambda_{\mathrm{guide}}(s)= \lambda_0\mathbb{I}[s<s_{\mathrm{switch}}]$, removing guidance thereafter to continue training with pure GRPO. Before $s_{\mathrm{switch}}$, RISE-RL adds $G$ privileged rollouts/evaluations per prompt and teacher-forces the retained trajectories, yielding a rollout/evaluation multiplier of $1+s_{\mathrm{switch}}/S$; wall-clock costs are reported in Appendix~\ref{app:compute_budget}.

\section{Experiments}

\begin{table*}[t]
\centering

\caption{Main results across four open-ended domains using
Qwen3-4B and Qwen3-14B. Each selected checkpoint is independently
evaluated three times, and the mean performance is reported.
Colored subscripts indicate absolute changes relative to the
corresponding Initial model, with improvements shown in green
and degradations in red. The best result within each model
scale is shown in bold.}
\label{tab:main_results}

\begingroup
\small
\setlength{\tabcolsep}{2.7pt}
\renewcommand{\arraystretch}{1.0}

\resizebox{\linewidth}{!}{%
\begin{tabular}{@{}lcccccccc@{\hspace{5pt}}c@{}}
\toprule

\multirow{2}{*}{\textbf{Method}}
& \multicolumn{2}{c}{\textbf{Writing}}
& \multicolumn{1}{c}{\textbf{Chat}}
& \multicolumn{3}{c}{\textbf{Health}}
& \multicolumn{2}{c}{\textbf{Science}}
& \multirow{2}{*}{\textbf{Avg.}}
\\

\cmidrule(lr){2-3}
\cmidrule(lr){4-4}
\cmidrule(lr){5-7}
\cmidrule(lr){8-9}

& \makecell{\textbf{Writing}\\[-1pt]\textbf{Bench}}
& \makecell{\textbf{Creative}\\[-1pt]\textbf{Writing-V3}}
& \makecell{\textbf{Arena-Hard}\\[-1pt]\textbf{V2}}
& \makecell{\textbf{Health}\\[-1pt]\textbf{Bench}}
& \makecell{\textbf{LLMEval}\\[-1pt]\textbf{-Med}}
& \textbf{MedQA}
& \textbf{GPQA}
& \makecell{\textbf{Research}\\[-1pt]\textbf{QA}}
&
\\

\midrule
\rowcolor{gray!18}
\multicolumn{10}{c}{\textit{Qwen3-4B (Non-Thinking)}} \\
\midrule

Initial
& \initres{56.14}
& \initres{40.28}
& \initres{7.91}
& \initres{37.43}
& \initres{64.76}
& \initres{64.79}
& \initres{42.34}
& \initres{64.29}
& \initres{47.24}
\\

SFT
& \gain{66.43}{10.29}
& \loss{32.96}{7.32}
& \gain{11.39}{3.48}
& \gain{40.73}{3.30}
& \gain{64.87}{0.11}
& \loss{59.77}{5.02}
& \loss{36.95}{5.39}
& \gain{69.19}{4.90}
& \gain{47.79}{0.55}
\\

OPD
& \loss{54.55}{1.59}
& \loss{27.78}{12.50}
& \loss{6.32}{1.59}
& \loss{35.46}{1.97}
& \loss{63.07}{1.69}
& \loss{61.63}{3.16}
& \loss{40.24}{2.10}
& \loss{60.35}{3.94}
& \loss{43.68}{3.56}
\\

Rubric-RL
& \gain{70.92}{14.78}
& \gain{43.30}{3.02}
& \gain{20.76}{12.85}
& \gain{52.00}{14.57}
& \gain{71.85}{7.09}
& \gain{68.44}{3.65}
& \gain{47.73}{5.39}
& \gain{74.52}{10.23}
& \gain{56.19}{8.95}
\\

RuscaRL
& \gain{56.75}{0.61}
& \loss{38.67}{1.61}
& \gain{9.41}{1.50}
& \gain{43.54}{6.11}
& \gain{65.81}{1.05}
& \loss{63.39}{1.40}
& \loss{41.41}{0.93}
& \gain{66.31}{2.02}
& \gain{48.16}{0.92}
\\

\rowcolor{oursbg}
\textbf{RISE-RL (Ours)}
& \gain{\textbf{72.15}}{16.01}
& \gain{\textbf{46.73}}{6.45}
& \gain{\textbf{20.89}}{12.98}
& \gain{\textbf{52.32}}{14.89}
& \gain{\textbf{72.22}}{7.46}
& \gain{\textbf{69.40}}{4.61}
& \gain{\textbf{49.12}}{6.78}
& \gain{\textbf{77.08}}{12.79}
& \gain{\textbf{57.49}}{10.25}
\\

\midrule
\rowcolor{gray!18}
\multicolumn{10}{c}{\textit{Qwen3-14B (Non-Thinking)}} \\
\midrule

Initial
& \initres{63.24}
& \initres{64.94}
& \initres{21.09}
& \initres{43.85}
& \initres{69.46}
& \initres{73.51}
& \initres{52.61}
& \initres{68.27}
& \initres{57.12}
\\

SFT
& \gain{73.31}{10.07}
& \loss{60.60}{4.34}
& \gain{33.74}{12.65}
& \gain{48.66}{4.81}
& \gain{71.63}{2.17}
& \gain{74.15}{0.64}
& \loss{39.90}{12.71}
& \gain{73.54}{5.27}
& \gain{59.44}{2.32}
\\

OPD
& \gain{66.05}{2.81}
& \loss{51.53}{13.41}
& \loss{20.07}{1.02}
& \gain{45.39}{1.54}
& \gain{71.56}{2.10}
& \gain{74.91}{1.40}
& \loss{48.82}{3.79}
& \gain{69.69}{1.42}
& \loss{56.00}{1.12}
\\

Rubric-RL
& \gain{76.89}{13.65}
& \loss{64.93}{0.01}
& \gain{49.94}{28.85}
& \gain{57.92}{14.07}
& \gain{76.40}{6.94}
& \gain{78.22}{4.71}
& \gain{55.98}{3.37}
& \gain{77.26}{8.99}
& \gain{67.19}{10.07}
\\

RuscaRL
& \loss{62.45}{0.79}
& \loss{42.78}{22.16}
& \gain{45.96}{24.87}
& \gain{50.92}{7.07}
& \gain{71.49}{2.03}
& \gain{77.39}{3.88}
& \loss{46.13}{6.48}
& \gain{74.10}{5.83}
& \gain{58.90}{1.78}
\\

\rowcolor{oursbg}
\textbf{RISE-RL (Ours)}
& \gain{\textbf{77.37}}{14.13}
& \gain{\textbf{70.96}}{6.02}
& \gain{\textbf{57.97}}{36.88}
& \gain{\textbf{59.44}}{15.59}
& \gain{\textbf{78.31}}{8.85}
& \gain{\textbf{81.57}}{8.06}
& \gain{\textbf{59.60}}{6.99}
& \gain{\textbf{78.96}}{10.69}
& \gain{\textbf{70.52}}{13.40}
\\

\bottomrule
\end{tabular}%
}
\endgroup

\end{table*}

\subsection{Experimental Setup}
\textbf{Datasets.}
To evaluate the effectiveness of our method across diverse open-ended domains, we construct the training corpus from RubricHub~\citep{li2026rubrichub}, a large-scale, multi-domain dataset equipped with fine-grained rubric-based evaluation criteria. RubricHub covers five domains: writing, health, chat, science, and instruction following. We select the writing, health, chat, and science subsets for reinforcement learning training and validation, thereby enabling a broad assessment of the proposed method across heterogeneous open-ended tasks.

\textbf{Benchmarks.} We evaluate RISE-RL across four domains using open-ended generation and verifiable reasoning tasks: (1) \textbf{Writing}: WritingBench~\citep{wu2026writingbench} and CreativeWriting-V3; (2) \textbf{Chat}: Arena-Hard-v2~\citep{li2024crowdsourced}; (3) \textbf{Health}: HealthBench~\citep{arora2025healthbench}, LLMEval-Med~\citep{zhang2025llmeval}, and MedQA~\citep{jin2021disease}; (4) \textbf{Science}: ResearchQA~\citep{yifei2026researchqa} and GPQA-Diamond~\citep{rein2023gpqa}. This suite assesses the model's ability to satisfy multidimensional rubrics in creative tasks and maintain factual rigor in knowledge-dense domains. All evaluations follow official protocols.

\textbf{Baselines.} We compare RISE-RL against imitation and RL-based baselines using Qwen3-4B/14B models:
(1) \textbf{SFT}: Fine-tuned on high-quality responses from RubricHub. 
(2) \textbf{OPD}~\citep{li2026rethinking}: On-policy distillation using Qwen3-14B as a teacher for the 4B model, and Qwen3-32B for the 14B model. 
(3) \textbf{Rubric-RL}~\citep{gunjal2025rubrics}: Standard RL using scalar rewards aggregated from fine-grained rubric criteria. 
(4) \textbf{RuscaRL}~\citep{zhou2025breaking}: A scaffold-mixed baseline that jointly optimizes rubric-conditioned and natural responses within a single group-relative objective.

\textbf{Implementation Details.}
We use \textbf{gpt-oss-120b} as the rubric grader, which has demonstrated high agreement with human judgment~\citep{li2026rubrichub}. SFT and OPD are implemented using SWIFT~\citep{zhao2025swift}, while all RL methods are implemented using verl~\citep{sheng2025hybridflow}. During the early guidance stage, RISE-RL additionally uses quality-filtered privileged candidates through a separate auxiliary objective. These candidates are excluded from the natural GRPO group and are no longer generated after $s_{\mathrm{switch}}$. All RL methods are trained separately for each domain.

For policy optimization, we use GRPO for the natural rollout branch. In RISE-RL, the shaping parameter is $\gamma=0.1$, and the guidance weight $\lambda_{\mathrm{guide}}$ is set to 0.05 for writing and chat, and 0.01 for health and science domains. Baselines, including RuscaRL, follow their best-reported configurations~\citep{zhou2025breaking}. Further implementation details, full training configurations, and evaluation protocols are provided in Appendix~\ref{app:implementation}.

\subsection{Main Results}
Table~\ref{tab:main_results} reports the downstream performance of RISE-RL. Overall, RISE-RL consistently achieves the strongest results across both Qwen3-4B and Qwen3-14B scales, effectively balancing complex alignment with autonomous exploration.

\textbf{Creative Domains: Enhancing Quality and Diversity.} In domains requiring high creativity and output diversity, such as \textbf{Writing} and \textbf{Chat}, RISE-RL shows substantial gains (up to \textbf{+6.0} and \textbf{+8.0} points on the 14B scale, respectively). On challenging benchmarks like \textbf{CreativeWriting-V3}, standard RL often struggles to discover high-reward trajectories or suffers from diversity collapse. RISE-RL's ability to target failed criteria exposes the model to rare creative modes (e.g., narrative coherence), while its \textit{guidance removal} mechanism ensures that the policy returns to unguided exploration, preserving the creative openness essential for these tasks.

\textbf{Knowledge-Dense Domains: Internalizing Logical Requirements.} In domains governed by strict factual and logical constraints, such as \textbf{Health} and \textbf{Science}, RISE-RL demonstrates that aligning with open-ended rubrics simultaneously reinforces performance on objective benchmarks. For instance, training on Health and Science subsets leads to a \textbf{+3.3} point gain on \textbf{MedQA} and a \textbf{+3.6} point gain on \textbf{GPQA-Diamond}. These results indicate that selective rubric guidance helps the model internalize domain-specific factual requirements and logical rigor, which are essential for both nuanced generation and verifiable reasoning.

\textbf{Comparison with Joint Optimization and Distillation.} Notably, RISE-RL maintains a substantial performance advantage over \textbf{OPD} and \textbf{RuscaRL}. OPD relies on teacher distillation, whose effectiveness depends on both the incremental value and the learnability of teacher trajectories; when the teacher--student gap is large, the additional supervision may be difficult to absorb, consistent with Rethinking OPD~\citep{li2026rethinking}. RuscaRL directly optimizes scaffold-conditioned samples, but the resulting behaviors do not always transfer reliably once the scaffold is removed. Consistent with the concerns raised in RGR-GRPO, we observe that this limitation varies across domains: in some cases, scaffold removal causes substantial performance degradation that subsequent on-policy training fails to fully recover~\citep{bi2025reward}. We provide further analysis of RuscaRL’s training dynamics in Appendix~\ref{subsec:appendix-ruscal}. In contrast, RISE-RL decouples privileged guidance from natural exploration, enabling more stable autonomous on-policy optimization.

\textbf{Compute-Budget Comparison.} Under the same hardware allocation, \textsc{RISE-RL} achieves a higher average score across WritingBench and CreativeWriting-V3 than Rubric-RL with approximately matched training time on Qwen3-4B (59.44 vs. 58.22; 21.48 vs. 21.08 hours). It also outperforms Rubric-RL with a matched rollout budget (59.44 vs. 57.68). Extending Rubric-RL training to 24.80 hours, exceeding RISE-RL in both training time and rollout count, does not close the gap. These results indicate that, within the examined setting and budget range, the gains cannot be attributed solely to additional training time or a larger rollout budget. Details are provided in Appendix~\ref{app:compute_budget}.

\subsection{Effect of Guidance Removal on Training Dynamics}
\begin{figure}[t]
\centering
\includegraphics[width=\textwidth]{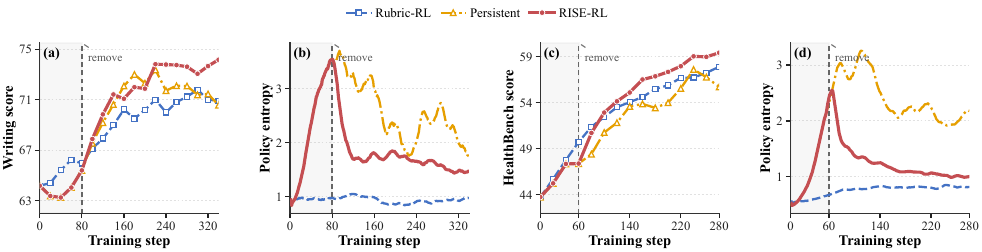} 
\caption{Training dynamics of Qwen3-14B on the writing and health domains.
We compare the benchmark performance and policy entropy of Rubric-RL,
persistent guidance, and RISE-RL on
(a--b) the writing domain and
(c--d) the health domain.}
\label{fig3}
\end{figure}

To analyze the role of dynamic guidance, we compare the training dynamics of RISE-RL with Rubric-RL and persistent guidance on \textbf{Writing} and \textbf{Health} domains (Fig.~\ref{fig3}).

\textbf{Training Dynamics.} Early guidance increases policy entropy from 0.9 to 3.5 in Writing and from 0.5 to 2.5 in Health, alongside performance gains. After guidance removal, entropy declines from its peak, while benchmark performance continues to improve. In contrast, persistent guidance maintains elevated, fluctuating entropy and exhibits late-stage performance plateaus or declines, including a 2.8-point drop from the peak Writing score. These observations support early guidance followed by autonomous on-policy optimization in the evaluated settings.

Guidance-removal sensitivity experiments further show that removal at steps 40 and 100 retains average improvements over Rubric-RL. This suggests some flexibility in removal timing; see Appendix~\ref{app:guidance_removal_timing} for details.
A reward-based automatic removal rule also achieves
comparable or higher average scores than manual scheduling on Qwen3-4B in Writing and Science (Appendix~\ref{app:automatic_guidance_removal}).

\subsection{Case Studies}
\begin{wrapfigure}[14]{r}{0.42\textwidth}
    \vspace{-0.8\baselineskip}
    \centering
    \includegraphics[
        width=\linewidth
    ]{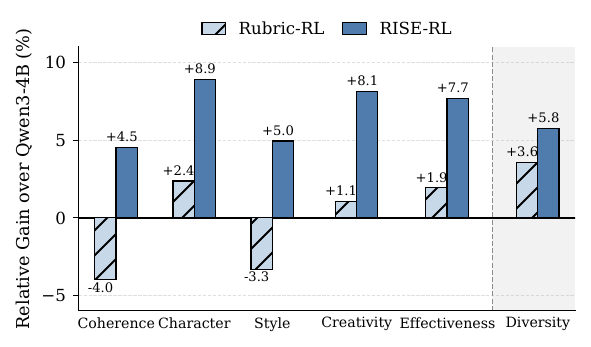}
    \caption{Relative gains over Qwen3-4B in overall creative-writing
    quality and output diversity.}
    \label{fig:creative_quality_diversity}
    \vspace{-0.5\baselineskip}
\end{wrapfigure}
\paragraph{Creative Writing: Quality and Diversity.}

As shown in Fig.~\ref{fig:creative_quality_diversity}, we report relative gains over Qwen3-4B in creative quality and output diversity. Quality is evaluated across five aggregated dimensions, with negative criteria reverse-scored before averaging. Diversity is assessed jointly over six responses per prompt across five set-level dimensions, using three random response permutations. All gains are computed as $(S_{\mathrm{method}}-S_{\mathrm{base}})/S_{\mathrm{base}}\times100\%$. RISE-RL improves all quality dimensions and achieves a \textbf{+5.8\%} diversity gain, whereas Rubric-RL degrades Coherence and Style. These results suggest that RISE-RL improves creative-writing quality while also enhancing output diversity. Details of the creative-writing evaluation and additional analyses are provided in Appendix~\ref{app:cw-evaluation}.

\textbf{Health: Acquisition of Persistent Criteria.}
To examine how RISE-RL resolves capability gaps, we analyze criterion-level failures on HealthBench (see Figure~\ref{health_case} in Appendix~\ref{app:health_case}). The initial policy exhibits persistent failures on critical requirements (e.g., \textit{Medical-advice disclaimer and pre-travel obstetric consultation}, 8/8 failures). While Rubric-RL struggles to recover these behaviors through scalar rewards, RISE-RL leverages selective guidance to reduce failures. By step 300, RISE-RL achieves zero failures on both the combined disclaimer-and-consultation criterion and the pregnancy-history criterion, whereas Rubric-RL still fails on these criteria in five and one of eight responses, respectively.

\subsection{Ablation Study}
\label{sec:choice-of-m}

\textbf{Sensitivity to Guidance Strength.}
We select $\lambda_{\mathrm{guide}}=0.05$ based on the largest training-reward increase among four coefficients in 140-step preliminary Writing runs with Qwen3-4B. A subsequent post-hoc evaluation also finds that this setting achieves the highest final CreativeWriting-V3 score among the tested values, providing additional support for the reward-based selection; see Appendix~\ref{app:guidance_strength} for details.

\begin{figure}[t]
\centering
\includegraphics[width=\textwidth]{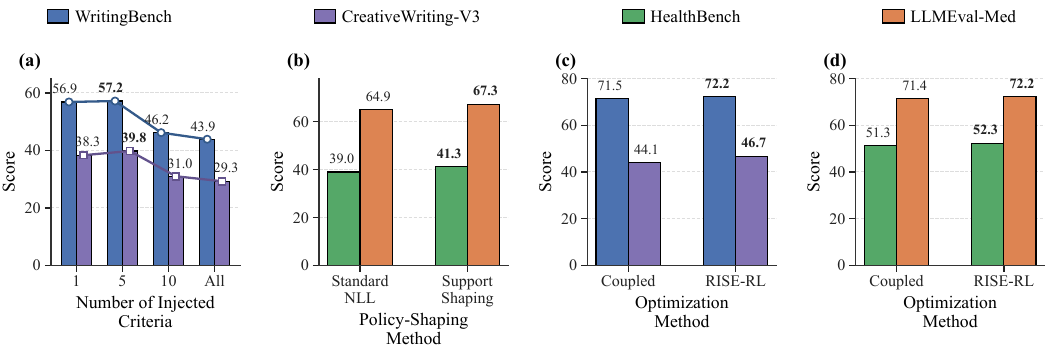} 
\caption{Ablations on Qwen3-4B. (a) Injected criterion count for Writing (“All”: all identified violations). (b) Policy support shaping for Health. (c–d) Coupled vs. decoupled optimization for Writing and Health, respectively. Panels (a–b) use matched short-run budgets.}
\label{ablation}
\end{figure}

\textbf{Effect of the Number of Injected Criteria.}
We vary the number of violated criteria injected into the privileged prompt over $\{1,5,10,\mathrm{All}\}$. All variants use Qwen3-4B in the Writing domain under identical settings (Figure~\ref{ablation}(a)). Injecting five criteria performs best on both WritingBench and CreativeWriting-V3, indicating a trade-off among robustness, coverage, and specificity. A single criterion may be overly sensitive to judge noise or an unrepresentative failure, whereas too many criteria reduce selectivity and introduce lower-priority, overlapping, or competing requirements that dilute the task signal. Overall, the results favor targeted over exhaustive criterion guidance.

\textbf{Effect of Policy Support Shaping.}
We evaluate the contribution of policy support shaping by replacing the token-level auxiliary loss $-\log[p_{j,t}/(p_{j,t}+\gamma)]$ with the standard NLL loss $-\log p_{j,t}$. The ablated variant retains online privileged-trajectory generation, reward filtering, trajectory-level gain weights, probability evaluation under the original prompt, and joint optimization with the natural GRPO branch. As shown in Figure~5(b), support shaping achieves higher scores on HealthBench and LLMEval-Med under the same guidance-loss coefficient and matched short-run budget. These results support the effectiveness of the policy support shaping component in the evaluated setting.

\textbf{Effect of Reward Filtering.}
We remove the positive-gain restriction and use all privileged trajectories with signed gains in the auxiliary loss. On Qwen3-4B in the Science domain, this reduces the average score across ResearchQA and GPQA-Diamond from 63.10 to 60.44, with decreases on both benchmarks. These results support restricting auxiliary learning to reward-improving trajectories in this setting. Details are provided in Appendix~\ref{app:reward_filtering}.

\textbf{Effect of Criterion Selection.}
We compare our criterion-selection strategy with two alternatives
on Qwen3-4B in Science: randomly selecting five criteria and
selecting the five criteria with the highest rubric weights.
Our strategy achieves an average score of 63.10 across
ResearchQA and GPQA-Diamond, compared with 59.53 and 60.46,
respectively.
These results support jointly considering criterion importance
and rollout failure frequency when selecting guidance.
Details are provided in Appendix~\ref{app:criterion_selection}.

\textbf{Coupled versus Decoupled Optimization.}
We compare our decoupled design with a coupled variant using the same criterion-selection and guidance-generation procedures. The coupled variant jointly optimizes privileged trajectories and natural rollouts within a shared group-relative objective, whereas \textsc{RISE-RL} incorporates privileged trajectories through a separate auxiliary objective. As shown in Figure~5(c--d), the decoupled design achieves higher scores on all four evaluated benchmarks across Writing and Health. These results support the effectiveness of the separate auxiliary-learning design relative to the evaluated coupled optimization variant.
\section{Conclusion and Discussion}
In this work, we introduced \textbf{RISE-RL}, a training paradigm that balances selective rubric-informed guidance with autonomous exploration for open-ended reinforcement learning. By identifying persistent criterion failures and providing decoupled, quality-filtered guidance during early training, RISE-RL effectively internalizes weakly supported behaviors from reward-improving guided trajectories. Extensive experiments across four diverse domains with 4B and 14B models demonstrate that RISE-RL consistently outperforms standard Rubric-RL and strong baselines while maintaining output diversity. Our results suggest that transitioning from targeted guidance to unguided exploration provides an effective approach for aligning large language models with complex, multidimensional human preferences.

Despite these gains, RISE-RL has several limitations. It introduces additional early-stage computation and depends on the coverage and reliability of explicit rubrics and their graders. Although preliminary automatic-removal results show that guidance can transition online based on normalized reward progress, the current shared threshold is empirically chosen and requires further validation across model scales and domains. More broadly, RISE-RL provides a natural starting point toward \textbf{recursive self-improvement (RSI)}. Future work could maintain a persistent feedback loop that identifies emerging capability gaps and adaptively adjusts guidance and the transition back to autonomous exploration, enabling continual \textbf{feedback–exploration–internalization} in open-ended environments.

\clearpage
\bibliographystyle{unsrtnat}
\bibliography{main}

\clearpage


\appendix

\startcontents[appendices]

\section*{Appendix}
\section*{Table of Contents}

\printcontents[appendices]{}{1}{%
  \setcounter{tocdepth}{2}%
}

\clearpage

\section{Dataset and Implementation Details}
\label{app:implementation}
\FloatBarrier

\subsection{Training Data}
\label{app:training_data}

We use the \textsc{RubricHub}~\citep{li2026rubrichub} dataset to construct the training data for both the reinforcement learning methods and the supervised fine-tuning baseline. We focus on four domains: \textsc{Writing}, \textsc{Chat}, \textsc{Health}, and \textsc{Science}, where \textsc{Health} corresponds to the \textsc{Medical} domain in the original \textsc{RubricHub} dataset. \textsc{RubricHub} provides two forms of training data used in our experiments: prompt--rubric pairs for reinforcement learning and high-quality prompt--response pairs for supervised fine-tuning.

\paragraph{Reinforcement Learning Data.}
For all reinforcement learning methods, including Rubric-RL, RuscaRL, and RISE-RL, we use the prompt--rubric pairs from the four selected \textsc{RubricHub} domains. We train each domain independently, resulting in a separate domain-specific model for Writing, Chat, Health, and Science. The complete reinforcement learning corpus contains 86,355 prompts, including 29,418 Science examples, 29,681 Health examples, 17,444 Writing examples, and 9,812 Chat examples. The detailed domain distribution is reported in Table~\ref{tab:rl_data_statistics}.

Each reinforcement learning instance consists of an open-ended prompt and a set of weighted, instance-specific evaluation criteria. For each generated response, \texttt{gpt-oss-120b} independently determines whether each rubric criterion is satisfied. The resulting criterion-level judgments are aggregated into a weight-normalized scalar reward for policy optimization. RISE-RL additionally uses these criterion-level judgments to identify repeatedly missed criteria and construct selective rubric guidance. No reference response is required during reinforcement learning.

\paragraph{Supervised Fine-Tuning Data.}
For the SFT baseline, we use the official supervised fine-tuning dataset released with \textsc{RubricHub}. It contains 26,194 high-quality prompt--response pairs mixed across the Writing, Chat, Health, and Science domains. The target responses are obtained through the multi-stage response refinement procedure used in \textsc{RubricHub}. We train the SFT baseline on the combined four-domain corpus using the standard next-token prediction objective.

The SFT and reinforcement learning datasets correspond to different data components provided by \textsc{RubricHub}. SFT uses a single mixed-domain corpus of refined responses, whereas each RL model is trained separately on the prompt--rubric pairs of one domain. SFT therefore serves as a supervised reference rather than a data-matched control. Our primary controlled comparison is between RISE-RL and Rubric-RL, which use the same backbone, domain-specific training data, rubric grader, and optimization configuration. The evaluation datasets are drawn from separate benchmark sources and are not used for training.

\begin{table}[t]
    \caption{Statistics of the reinforcement learning training data.}
    \label{tab:rl_data_statistics}
    \centering
    
    \small
    \setlength{\tabcolsep}{5pt}
    \renewcommand{\arraystretch}{0.95}
    \begin{tabular}{lrr}
        \toprule
        \textbf{Domain} & \textbf{Prompts} & \textbf{Share} \\
        \midrule
        Science & 29,418 & 34.07\% \\
        Health  & 29,681 & 34.37\% \\
        Writing & 17,444 & 20.20\% \\
        Chat    &  9,812 & 11.36\% \\
        \midrule
        \textbf{Total} & \textbf{86,355} & \textbf{100.00\%} \\
        \bottomrule
    \end{tabular}
\end{table}

\begin{table}[t]
    \centering
    \caption{Summary of the training data used in our experiments.}
    \label{tab:training_data_summary}
    \small
    \setlength{\tabcolsep}{5pt}
    \renewcommand{\arraystretch}{0.95}
    \begin{tabular}{lcc}
        \toprule
        \textbf{Training Paradigm} &
        \textbf{Domain Organization} &
        \textbf{Examples (Aggregate)} \\
        \midrule
        SFT & One mixed-domain run & 26,194 \\        
        RL & Four separate domain-specific runs & 86,355 \\        \bottomrule
    \end{tabular}
\end{table}

\subsection{RISE-RL Training Procedure}

Algorithm~\ref{alg:rise} summarizes the complete RISE-RL training
procedure, including natural-prompt policy optimization, selective
privileged-trajectory generation, quality filtering, policy-support
shaping, and the transition to guidance-free training.

\begin{algorithm}[t]
\caption{Training Procedure of RISE-RL}
\label{alg:rise}
\small
\begin{algorithmic}[1]

\Require Training set $\mathcal{D}$, policy $\pi_\theta$, total training
steps $S$, group size $G$, feedback budget $M$, guidance weight
$\lambda_0$, gain cap $A_{\max}$, and removal step
$s_{\mathrm{switch}}$

\For{$s=1,\ldots,S$}

    \State Sample $q\sim\mathcal{D}$ and generate natural rollouts
    \[
    \mathcal{O}^{\mathrm{nat}}
    \leftarrow
    \mathrm{Rollout}(\pi_{\theta_{\mathrm{old}}},q,G)
    \]

    \State Evaluate the natural rollouts:
    \[
    (Z^{\mathrm{nat}},R^{\mathrm{nat}})
    \leftarrow
    \mathrm{Evaluate}
    (\mathcal{O}^{\mathrm{nat}},\mathcal{C}(q))
    \]

    \State Compute the GRPO loss:
    \[
    \mathcal{L}_{\mathrm{GRPO}}
    \leftarrow
    \mathrm{GRPO}
    (\mathcal{O}^{\mathrm{nat}},R^{\mathrm{nat}})
    \]

    \State Set
    $\mathcal{L}\leftarrow\mathcal{L}_{\mathrm{GRPO}}$

    \If{$s<s_{\mathrm{switch}}$}

        \State Select up to $M$ positive-priority criteria
        $\mathcal{C}^{\mathrm{fb}}$ from $Z^{\mathrm{nat}}$;
        if none exists, skip to the policy update

        \State Construct the privileged prompt
        \[
        q^{\mathrm{priv}}
        \leftarrow
        q\oplus\mathcal{C}^{\mathrm{fb}}
        \]
        and generate privileged rollouts
        $\mathcal{O}^{\mathrm{priv}}$

        \State Evaluate $\mathcal{O}^{\mathrm{priv}}$
        using the complete rubric $\mathcal{C}(q)$

        \State Retain trajectories $\mathcal{O}^{\mathrm{ref}}$
        with positive normalized reward gains

        \State Teacher-force $\mathcal{O}^{\mathrm{ref}}$
        under the original prompt $q$ to estimate
        natural-policy token support $p_{j,t}$

        \State Compute policy-support weights, clipped normalized trajectory gains, and
        $\mathcal{L}_{\mathrm{guide}}$

        \State Update the total loss:
        \[
        \mathcal{L}
        \leftarrow
        \mathcal{L}_{\mathrm{GRPO}}
        +\lambda_0\mathcal{L}_{\mathrm{guide}}
        \]

    \EndIf

    \State Update the policy parameters:
    \[
    \theta
    \leftarrow
    \mathrm{Update}(\theta,\mathcal{L})
    \]

\EndFor

\State \Return $\pi_\theta$

\end{algorithmic}
\end{algorithm}

\subsection{Training Configuration}
\label{app:training_configuration}


\paragraph{Rubric-RL and RISE-RL.} We implement Rubric-RL and RISE-RL using the \texttt{verl} framework.
For a controlled comparison, both methods use the same backbone model,
training data, sampling configuration and optimization hyperparameters. RISE-RL additionally introduces a selective
guidance branch with several method-specific hyperparameters. The complete
training configuration is summarized in
Table~\ref{tab:rl_training_configuration}.

\begin{table*}[t]
    \caption{Training configurations for Rubric-RL and RISE-RL.}
    \label{tab:rl_training_configuration}
    \centering
    \small
    \setlength{\tabcolsep}{6pt}
    \renewcommand{\arraystretch}{1.05}
    \begin{tabular}{ll}
        \toprule
        \textbf{Category} & \textbf{Configuration} \\
        \midrule

        \multirow{14}{*}{Shared}
        & Training batch size: \texttt{64} \\
        & Maximum prompt length: \texttt{4096} \\
        & Maximum response length: \texttt{8192} \\
        & Number of rollouts per prompt: \texttt{8} \\
        & Overlong length: \texttt{4096} \\
        & Overlong penalty factor: \texttt{0.5} \\
        & Learning rate: \texttt{1e-6} \\
        & Warmup steps: \texttt{10} \\
        & Weight decay: \texttt{0.1} \\
        & Entropy coefficient: \texttt{0.0} \\
        & Reward range: \texttt{[0,1]} \\
        & KL-loss coefficient: \texttt{0.0} \\
        & Asymmetric PPO clipping $\epsilon_{\mathrm{low}}$: \texttt{0.2} \\
        & Asymmetric PPO clipping $\epsilon_{\mathrm{high}}$: \texttt{0.28} \\
        \midrule

        \multirow{4}{*}{RISE-RL}
        & Number of guided re-rollouts: \texttt{8} \\
        & Guidance support coefficient $\gamma$: \texttt{0.1} \\
        & Guidance-loss coefficient $\lambda_{\mathrm{guide}}$: \texttt{0.05 (Writing/Chat), 0.01 (Health/Science)} \\
        & Number of injected criteria $M$: \texttt{5} \\
        & Trajectory-gain cap $A_{\max}$: \texttt{5} \\
        \midrule

        Hardware
        & 8 $\times$ NVIDIA H200 GPUs \\
        \bottomrule
    \end{tabular}
\end{table*}

\paragraph{RuscaRL.}
We implement RuscaRL~\citep{zhou2025breaking} following its original scaffolding strategy and
training configuration. Specifically, RuscaRL
applies linear intra-group scaffolding differentiation and gradually
reduces the rubric scaffolding using a step-wise sigmoid schedule.
The main training configuration is summarized in
Table~\ref{tab:ruscarl_training_configuration}. We train RuscaRL for 500 optimization steps in all domains.

\paragraph{Checkpoint and Guidance Schedule Selection.} For \textsc{RISE-RL}, Rubric-RL, and our reproduction of RuscaRL, checkpoints are selected based on training reward convergence. Downstream benchmark scores are not used for checkpoint selection. For the main \textsc{RISE-RL} experiments, the early guidance-removal step is determined once for each model--domain setting based on the observed training-reward dynamics, rather than selected through a downstream-benchmark sweep over removal steps. Timing sensitivity and an automatic reward-based alternative are evaluated in Appendices~\ref{app:guidance_removal_timing} and~\ref{app:automatic_guidance_removal}, respectively. Separately, the Writing guidance coefficient is selected based on the largest increase in training reward over 140-step preliminary runs on Qwen3-4B and then fixed for the full Writing training runs at both model scales. The subsequent CreativeWriting-V3 evaluation in Appendix~\ref{app:guidance_strength} is a post-hoc sensitivity analysis and is not used to select or readjust the coefficient.

\paragraph{Representative Removal Schedules.}
We illustrate the schedule selection using the Qwen3-14B Writing and Health settings shown in Figure~\ref{fig3}. Based only on preliminary training-reward dynamics, we set $s_{\mathrm{switch}}=80$ for Writing and $s_{\mathrm{switch}}=60$ for Health, near the onset of a plateau in the smoothed privileged--natural reward gap. These values were fixed before downstream evaluation. The vertical ``remove'' markers in Figure~\ref{fig3} indicate the corresponding transitions. After guidance removal, policy entropy decreases from its guidance-induced peak while benchmark performance continues to improve, suggesting that the acquired behaviors are consolidated through autonomous on-policy optimization. In contrast, persistent guidance maintains elevated or fluctuating entropy and leads to late-stage performance plateaus or degradation. These observations motivate removing guidance once its marginal training benefit diminishes.


\begin{table*}[t]
    \caption{Training configuration for RuscaRL.}
    \label{tab:ruscarl_training_configuration}
    \centering
    \small
    \setlength{\tabcolsep}{6pt}
    \renewcommand{\arraystretch}{1.05}
    \begin{tabular}{ll}
        \toprule
        \textbf{Category} & \textbf{Configuration} \\
        \midrule

        \multirow{4}{*}{RuscaRL}
        & RL algorithm: GRPO \\
        & Inter-step scaffolding decay:
          step sigmoid ($\alpha=125$, $t_0=0.2$) \\
        & Intra-group scaffolding differentiation: linear \\
        & Rubric grader: \texttt{gpt-oss-120b} \\
        \midrule

        \multirow{4}{*}{Sampling}
        & Temperature: 0.7 \\
        & Top-$p$: 0.8; Top-$k$: 20 \\
        & Rollouts per prompt: 8 \\
        & Maximum response length: 8192 \\
        \midrule

        \multirow{7}{*}{Training}
        & Optimizer: Adam \\
        & Learning rate: $1\times10^{-6}$ (constant) \\
        & Training batch size: 64 \\
        & Mini-batch size: 32 \\
        & KL-loss coefficient: $1\times10^{-3}$ \\
        & Entropy coefficient: 0 \\
        & Total training steps: 500 \\
        \midrule

        Hardware
        & 8 $\times$ NVIDIA H200 GPUs \\
        \bottomrule
    \end{tabular}
\end{table*}

\paragraph{Supervised Fine-Tuning.}
We implement the SFT baseline using \texttt{SWIFT} with full-parameter fine-tuning. The model is trained on the mixed-domain RubricHub SFT corpus for three epochs using bfloat16 precision and DeepSpeed ZeRO-3, and we evaluate the final checkpoint. The main training hyperparameters are summarized in Table~\ref{tab:sft_training_configuration}.

\begin{table*}[t]
    \caption{Training configuration for the SFT baseline.}
    \label{tab:sft_training_configuration}
    \centering
    \small
    \setlength{\tabcolsep}{6pt}
    \renewcommand{\arraystretch}{1.05}
    \begin{tabular}{ll}
        \toprule
        \textbf{Hyperparameter} & \textbf{Configuration} \\
        \midrule
        Training framework & \texttt{SWIFT} \\
        Fine-tuning type & Full-parameter fine-tuning \\
        Training epochs & 3 \\
        Learning rate & $1\times10^{-5}$ \\
        Per-device batch size & 2 \\
        Gradient accumulation steps & 4 \\
        Maximum sequence length & 20,000 \\
        Warmup ratio & 0.05 \\
        Validation split & 1\% \\
        Precision & bfloat16 \\
        Distributed training & DeepSpeed ZeRO-3 \\
        Hardware & 8 $\times$ NVIDIA H200 GPUs \\
        \bottomrule
    \end{tabular}
\end{table*}

\paragraph{On-Policy Distillation.}
We implement OPD~\citep{li2026rethinking} using the \texttt{SWIFT} framework with full-parameter
fine-tuning. Following the standard on-policy distillation setup, the student
model generates trajectories from its current policy, while a larger teacher
model provides token-level supervision on the student-generated states. For
the 4B student, we use Qwen3-14B as the teacher; for the 14B student, we use
Qwen3-32B as the teacher. All models are trained with bfloat16 precision and
DeepSpeed ZeRO-2. The main training hyperparameters are summarized in
Table~\ref{tab:opd_training_configuration}.

\begin{table*}[t]
    \caption{Training configuration for the OPD baseline.}
    \label{tab:opd_training_configuration}
    \centering
    \small
    \setlength{\tabcolsep}{6pt}
    \renewcommand{\arraystretch}{1.05}
    \begin{tabular}{ll}
        \toprule
        \textbf{Hyperparameter} & \textbf{Configuration} \\
        \midrule
        Training framework & \texttt{SWIFT} \\
        Training objective & On-policy distillation (GKD) \\
        Student--teacher pairs
        & 4B--14B; 14B--32B \\
        Fine-tuning type & Full-parameter fine-tuning \\
        Training steps & 50 \\
        Learning rate & $1\times10^{-5}$ \\
        Per-device batch size & 4 \\
        Gradient accumulation steps & 16 \\
        Maximum sequence length & 16,000 \\
        Maximum completion length & 8,192 \\
        Warmup ratio & 0.05 \\
        Precision & bfloat16 \\
        Distributed training & DeepSpeed ZeRO-2 \\
        Hardware & 8 $\times$ NVIDIA H200 GPUs \\
        \bottomrule
    \end{tabular}
\end{table*}

\subsection{Evaluation Implementations}
\label{app:evaluation_implementations}

We evaluate all models on eight benchmarks spanning four domains:
Writing, Chat, Health, and Science. The evaluation suite includes both
open-ended generation tasks and multiple-choice question answering
tasks. For open-ended benchmarks, we follow the official rubric-based
or pairwise-comparison protocols. For multiple-choice benchmarks, we
report answer accuracy based on the extracted final choice. Unless
otherwise specified, evaluation is performed without rubric guidance
or any other privileged information.We evaluate on the complete official test sets without subsampling
and follow the official evaluation implementations and default
generation settings.

For each selected checkpoint, we independently repeat the
complete evaluation pipeline three times, including response
generation and LLM-based judging, and report the mean and
standard deviation. This protocol quantifies the end-to-end
evaluation variability for a fixed checkpoint. The question types and evaluation metrics are summarized in
Table~\ref{tab:evaluation_summary}.

\paragraph{Writing.}
We evaluate writing ability using WritingBench and CreativeWriting-V3.

\textbf{WritingBench}~\citep{wu2026writingbench} evaluates long-form responses to diverse
real-world writing instructions using instance-specific criteria
covering multiple dimensions of writing quality. We follow the official
rubric-based evaluation protocol and report the aggregated rubric
score.

\textbf{CreativeWriting-V3} evaluates open-ended creative writing,
including narrative coherence, characterization, style, originality,
and overall effectiveness. We follow the official rubric-based
evaluation protocol and report the aggregate rubric score.

\paragraph{Chat.}
We use Arena-Hard-v2~\citep{li2024crowdsourced} to evaluate general open-ended
instruction-following and conversational ability. Candidate responses
are compared against reference-model responses under the official
style-controlled pairwise evaluation protocol, and we report the
resulting win rate.

\paragraph{Health.}
We evaluate health-domain capabilities on HealthBench~\citep{arora2025healthbench}, LLMEval-Med~\citep{zhang2025llmeval},
and MedQA~\citep{jin2021disease}.

\textbf{HealthBench} consists of open-ended healthcare conversations
evaluated using physician-written, conversation-specific rubric
criteria with importance weights. The final score is computed from the
weighted satisfaction of these criteria.

\textbf{LLMEval-Med} evaluates open-ended responses to real-world
clinical questions using expert-written reference answers and
checklists. We follow the official checklist-based evaluation pipeline
and report the overall usability score.

\textbf{MedQA} is a multiple-choice medical question-answering
benchmark. We report exact-match accuracy of the predicted answer
choice.

\paragraph{Science.}
We evaluate scientific knowledge and reasoning using ResearchQA~\citep{yifei2026researchqa} and
GPQA-Diamond~\citep{rein2023gpqa}.

\textbf{ResearchQA} evaluates long-form scholarly question answering
using query-specific rubric items. We report the normalized
rubric-coverage score.

\textbf{GPQA-Diamond} is a graduate-level multiple-choice benchmark
covering biology, physics, and chemistry. We report exact-match
accuracy of the predicted answer choice.

\begin{table*}[t]
    \caption{Summary of the evaluation protocols.}
    \label{tab:evaluation_summary}
    \centering
    \small
    \setlength{\tabcolsep}{5pt}
    \renewcommand{\arraystretch}{1.05}
    \begin{tabular}{llll}
        \toprule
        \textbf{Domain} &
        \textbf{Benchmark} &
        \textbf{Question Type} &
        \textbf{Metric} \\
        \midrule

        \multirow{2}{*}{Writing}
        & WritingBench
        & Open-ended writing
        & Rubric score \\
        & CreativeWriting-V3
        & Creative writing
        & Rubric score \\
        \midrule

        Chat
        & Arena-Hard-v2
        & Open-ended dialogue
        & Style-controlled win rate \\
        \midrule

        \multirow{3}{*}{Health}
        & HealthBench
        & Open-ended healthcare
        & Weighted rubric score \\
        & LLMEval-Med
        & Open-ended clinical QA
        & Overall usability \\
        & MedQA
        & Multiple choice
        & Exact-match accuracy \\
        \midrule

        \multirow{2}{*}{Science}
        & ResearchQA
        & Long-form scholarly QA
        & Rubric coverage \\
        & GPQA-Diamond
        & Multiple choice
        & Exact-match accuracy \\
        \bottomrule
    \end{tabular}
\end{table*}

\section{Additional Experimental Results}
\label{app:additional_results}

\subsection{Main Results with Standard Deviations}
\label{subsec:appendix-result-with-deviations}

Table~\ref{tab:main_results_with_standard_deviations} reports the full benchmark results with standard deviations. For each method and model scale, we independently evaluate the same selected checkpoint three times and report the mean and standard deviation across these evaluations. These standard deviations quantify variability from repeated response generation and evaluation for a fixed checkpoint.

Table~\ref{tab:main_results_with_standard_deviations} indicates generally limited variability across the three independent evaluations: 62 of the 96 benchmark entries have standard deviations at or below 0.5 points. Evaluation variability differs across benchmarks, with WritingBench and HealthBench consistently exhibiting standard deviations below 0.5 points across all methods and both model scales. GPQA exhibits the largest standard deviation in every method--scale setting, ranging from 1.09 to 2.92 points. This greater variability may partly reflect its relatively small evaluation set, which makes the aggregate score more sensitive to changes in individual predictions. Overall, these results indicate that most reported benchmark scores are reasonably consistent across repeated evaluations, with greater variability observed on GPQA.

\begin{table*}[t]
\caption{Full results corresponding to Table~\ref{tab:main_results}, with evaluation standard deviations, for Qwen3-4B and Qwen3-14B across four domains. Each benchmark entry reports the mean $\pm$ standard deviation over three independent evaluations of the same selected checkpoint. Avg.\ denotes the unweighted average of the eight benchmark means. The highest mean score for each benchmark and the highest Avg.\ within each model scale are shown in bold.}
\label{tab:main_results_with_standard_deviations}
\centering

{\small
\setlength{\tabcolsep}{3.0pt}

\resizebox{\textwidth}{!}{%
\begin{tabular}{@{}lcccccccc@{\hspace{5pt}}c@{}}
\toprule

\multirow{2}{*}{\textbf{Method}}
&
\multicolumn{2}{c}{\textbf{Writing}}
&
\multicolumn{1}{c}{\textbf{Chat}}
&
\multicolumn{3}{c}{\textbf{Health}}
&
\multicolumn{2}{c}{\textbf{Science}}
&
\multirow{2}{*}{\textbf{Avg.}}
\\

\cmidrule(lr){2-3}
\cmidrule(lr){4-4}
\cmidrule(lr){5-7}
\cmidrule(lr){8-9}

&
\makecell{\textbf{Writing}\\[-1pt]\textbf{Bench}}
&
\makecell{\textbf{Creative}\\[-1pt]\textbf{Writing-V3}}
&
\makecell{\textbf{Arena-Hard}\\[-1pt]\textbf{V2}}
&
\makecell{\textbf{Health}\\[-1pt]\textbf{Bench}}
&
\makecell{\textbf{LLMEval}\\[-1pt]\textbf{-Med}}
&
\textbf{MedQA}
&
\textbf{GPQA}
&
\makecell{\textbf{Research}\\[-1pt]\textbf{QA}}
&
\\

\midrule
\multicolumn{10}{c}{\textit{Qwen3-4B (Non-Thinking)}} \\
\midrule

Initial
& \res{56.14}{0.10}
& \res{40.28}{0.77}
& \res{7.91}{0.07}
& \res{37.43}{0.12}
& \res{64.76}{0.28}
& \res{64.79}{0.12}
& \res{42.34}{1.98}
& \res{64.29}{0.22}
& 47.24
\\

SFT
& \res{66.43}{0.30}
& \res{32.96}{1.00}
& \res{11.39}{0.13}
& \res{40.73}{0.11}
& \res{64.87}{0.82}
& \res{59.77}{0.85}
& \res{36.95}{2.68}
& \res{69.19}{0.43}
& 47.79
\\

OPD
& \res{54.55}{0.22}
& \res{27.78}{0.52}
& \res{6.32}{0.11}
& \res{35.46}{0.19}
& \res{63.07}{0.15}
& \res{61.63}{0.17}
& \res{40.24}{2.01}
& \res{60.35}{0.36}
& 43.68
\\

Rubric-RL
& \res{70.92}{0.21}
& \res{43.30}{0.36}
& \res{20.76}{0.62}
& \res{52.00}{0.01}
& \res{71.85}{0.34}
& \res{68.44}{0.39}
& \res{47.73}{1.20}
& \res{74.52}{0.71}
& 56.19
\\

RuscaRL
& \res{56.75}{0.39}
& \res{38.67}{0.52}
& \res{9.41}{0.49}
& \res{43.54}{0.13}
& \res{65.81}{0.50}
& \res{63.39}{0.36}
& \res{41.41}{2.92}
& \res{66.31}{0.92}
& 48.16
\\

\textbf{RISE-RL (Ours)}
& \best{72.15}{0.12}
& \best{46.73}{0.56}
& \best{20.89}{0.25}
& \best{52.32}{0.25}
& \best{72.22}{0.73}
& \best{69.40}{0.25}
& \best{49.12}{2.52}
& \best{77.08}{0.29}
& \textbf{57.49}
\\

\midrule
\multicolumn{10}{c}{\textit{Qwen3-14B (Non-Thinking)}} \\
\midrule

Initial
& \res{63.24}{0.15}
& \res{64.94}{0.40}
& \res{21.09}{0.31}
& \res{43.85}{0.22}
& \res{69.46}{0.52}
& \res{73.51}{0.44}
& \res{52.61}{1.98}
& \res{68.27}{0.27}
& 57.12
\\

SFT
& \res{73.31}{0.12}
& \res{60.60}{0.88}
& \res{33.74}{0.39}
& \res{48.66}{0.43}
& \res{71.63}{0.39}
& \res{74.15}{0.98}
& \res{39.90}{1.69}
& \res{73.54}{0.32}
& 59.44
\\

OPD
& \res{66.05}{0.08}
& \res{51.53}{0.60}
& \res{20.07}{0.29}
& \res{45.39}{0.18}
& \res{71.56}{0.35}
& \res{74.91}{1.12}
& \res{48.82}{2.43}
& \res{69.69}{0.33}
& 56.00
\\

Rubric-RL
& \res{76.89}{0.20}
& \res{64.93}{0.71}
& \res{49.94}{0.42}
& \res{57.92}{0.06}
& \res{76.40}{0.41}
& \res{78.22}{0.68}
& \res{55.98}{2.05}
& \res{77.26}{0.08}
& 67.19
\\

RuscaRL
& \res{62.45}{0.13}
& \res{42.78}{1.08}
& \res{45.96}{0.48}
& \res{50.92}{0.17}
& \res{71.49}{0.65}
& \res{77.39}{0.49}
& \res{46.13}{2.74}
& \res{74.10}{0.34}
& 58.90
\\

\textbf{RISE-RL (Ours)}
& \best{77.37}{0.10}
& \best{70.96}{0.53}
& \best{57.97}{0.31}
& \best{59.44}{0.11}
& \best{78.31}{0.20}
& \best{81.57}{0.91}
& \best{59.60}{1.09}
& \best{78.96}{0.15}
& \textbf{70.52}
\\

\bottomrule
\end{tabular}%
}
}
\end{table*}


\subsection{Compute-Budget Comparison}
\label{app:compute_budget}
We compare RISE-RL and Rubric-RL using Qwen3-4B in the Writing domain under the same hardware allocation. Rubric-RL requires approximately 3.72 minutes per optimization step. RISE-RL incurs additional computation only during its initial guidance phase, which lasts for 80 steps in this setting and requires approximately 6.87 minutes per step because of the additional privileged rollouts, complete-rubric evaluation, and teacher-forcing computation. After guidance is removed, RISE-RL follows the same natural-prompt training procedure as Rubric-RL and therefore has approximately the same per-step cost.

We consider complementary controls for wall-clock time and rollout budget. First, we extend Rubric-RL to 340 steps, resulting in a training time approximately matched to that of RISE-RL (21.08 versus 21.48 hours). Second, we train Rubric-RL for 360 steps to match the total number of generated and rubric-evaluated trajectories. 
Specifically, with a training batch size of $B=64$ and a rollout group
size of $G=8$, RISE-RL generates
\[
280 \times B \times G + 80 \times B \times G = 184{,}320
\]
trajectories: $280 \times B \times G$ natural trajectories over the
full training run and an additional $80 \times B \times G$ privileged
trajectories during the first 80 steps. Rubric-RL at 360 steps generates
\[
360 \times B \times G = 184{,}320
\]
natural trajectories, exactly matching this rollout budget. We additionally report Rubric-RL at 400 steps, which exceeds RISE-RL in both wall-clock time and the number of generated trajectories.

As shown in Table~\ref{tab:writing_compute_budget}, RISE-RL achieves an average score of 59.44 across WritingBench and CreativeWriting-V3, compared with 58.22 for the approximately time-matched 340-step Rubric-RL run and 57.68 for the rollout-matched 360-step run. Extending Rubric-RL to 400 steps further increases its training budget but does not improve its average performance. RISE-RL also remains better on both individual benchmarks under all three extended Rubric-RL budgets. These results indicate that, within the examined Qwen3-4B Writing setting, the gains of RISE-RL cannot be explained solely by longer training time or a larger number of generated and reward-evaluated trajectories.

\begin{table}[t]
    \centering
    \caption{
       Compute-budget comparison on Qwen3-4B in Writing. The 340-step Rubric-RL run approximately matches the wall-clock training time of RISE-RL, while the 360-step run matches its total number of generated and rubric-evaluated trajectories. The 400-step run exceeds RISE-RL in both budgets. Avg.\ denotes the unweighted mean of the two benchmark scores. Training times are approximate.}
    \label{tab:writing_compute_budget}
    \small
    \renewcommand{\arraystretch}{1.15}
    \setlength{\tabcolsep}{5pt}

    \begin{tabular*}{\linewidth}{
        @{\extracolsep{\fill}} lccccc @{}
    }
        \toprule
        & \multicolumn{2}{c}{Training budget}
        & \multicolumn{3}{c}{Writing performance} \\
        \cmidrule(lr){2-3}
        \cmidrule(l){4-6}
        Method
        & Steps
        & Time (h)
        & WritingBench
        & CreativeWriting-V3
        & Avg. \\
        \midrule
        Rubric-RL
        & 280 & 17.36 & 70.92 & 43.30 & 57.11 \\
        Rubric-RL
        & 340 & 21.08 & 71.86 & 44.57 & 58.22 \\
        Rubric-RL
        & 360 & 22.32 & 71.38 & 43.97 & 57.68 \\
        Rubric-RL
        & 400 & 24.80 & 71.92 & 42.53 & 57.23 \\
        \midrule
        \textsc{RISE-RL} (Ours)
        & 280 & 21.48
        & \textbf{72.15}
        & \textbf{46.73}
        & \textbf{59.44} \\
        \bottomrule
    \end{tabular*}
\end{table}

\subsection{Effect of Reward Filtering}
\label{app:reward_filtering}

Privileged guidance targets selected rubric criteria, but improvement on these criteria does not necessarily imply a higher reward under the complete rubric. \textsc{RISE-RL} therefore retains only guided trajectories with positive normalized reward gains relative to the natural-rollout mean, $A_j^{\mathrm{ref}}
=
\frac{r_j^{\mathrm{priv}}-\bar{r}^{\mathrm{nat}}}
{\sigma^{\mathrm{nat}}+\epsilon_A}
>0$. To evaluate this mechanism, we compare \textsc{RISE-RL} with a no-filter variant on Qwen3-4B in the Science domain. The no-filter variant uses all privileged trajectories in the auxiliary loss and its normalization, using symmetrically clipped weights $\widetilde A_j = \operatorname{clip}\!\left(A_j^{\mathrm{ref}}, -A_{\max}, A_{\max}\right)$ with $A_{\max}=5$, so that negative normalized gains retain their signs while their magnitudes remain bounded. We measure overall Science performance using the unweighted mean of ResearchQA and GPQA-Diamond scores.

As shown in Table~\ref{tab:reward_filtering}, removing reward filtering lowers ResearchQA and GPQA-Diamond scores by 1.15 and 4.17 points, respectively, reducing their average from 63.10 to 60.44. These results favor restricting auxiliary learning to reward-improving trajectories over using all privileged trajectories with signed normalized gains in this setting.

\begin{table}[t]
    \centering
    \caption{
        Reward-filtering ablation on Qwen3-4B in the
        Science domain. Avg.\ denotes the unweighted
        mean of ResearchQA and GPQA-Diamond scores.
        Best scores are bold.
    }
    \label{tab:reward_filtering}
    \small
    \renewcommand{\arraystretch}{1.15}
    \setlength{\tabcolsep}{10pt}

    \begin{tabular}{lccc}
        \toprule
        Method & ResearchQA & GPQA-Diamond & Avg. \\
        \midrule
        \textsc{RISE-RL} w/o reward filtering
        & 75.93 & 44.95 & 60.44 \\
        \textsc{RISE-RL}
        & \textbf{77.08}
        & \textbf{49.12}
        & \textbf{63.10} \\
        \bottomrule
    \end{tabular}
\end{table}

\subsection{Effect of Criterion Selection}
\label{app:criterion_selection}

We examine whether the criterion-selection strategy contributes
to the effectiveness of \textsc{RISE-RL} using Qwen3-4B
in the Science domain.
We compare the default strategy with two alternatives:
\textit{Random}, which randomly selects five rubric criteria,
and \textit{Weight-only}, which selects the five criteria
with the highest rubric importance weights $w_k$.
The default strategy also selects five criteria, but ranks
them by
\[
    p_k = w_k
    \left(1 - \frac{1}{G}\sum_{i=1}^{G} z_{i,k}\right),
\]
where $z_{i,k}$ indicates whether natural rollout $i$
satisfies criterion $k$.
This priority combines criterion importance with the frequency
of failures across natural rollouts.
The three variants therefore use the same number of
injected criteria.

As shown in Table~\ref{tab:criterion_selection},
the default strategy achieves an average score of 63.10,
exceeding Random and Weight-only by 3.57 and 2.64 points,
respectively.
It also achieves higher scores on both individual benchmarks.
These results support using rollout-dependent failure
information together with rubric importance to select
guidance in this setting.

\begin{table}[t]
    \centering
    \caption{
        Criterion-selection ablation on Qwen3-4B in Science.
        Each strategy selects five criteria.
        Avg.\ denotes the unweighted mean of ResearchQA
        and GPQA-Diamond scores. Best scores are bold.
    }
    \label{tab:criterion_selection}
    \small
    \renewcommand{\arraystretch}{1.15}
    \setlength{\tabcolsep}{10pt}

    \begin{tabular}{lccc}
        \toprule
        Selection strategy
        & ResearchQA & GPQA-Diamond & Avg. \\
        \midrule
        Random
        & 76.13 & 42.93 & 59.53 \\
        Weight-only
        & 75.97 & 44.95 & 60.46 \\
        \midrule
        Default (\textsc{RISE-RL})
        & \textbf{77.08}
        & \textbf{49.12}
        & \textbf{63.10} \\
        \bottomrule
    \end{tabular}
\end{table}

\subsection{Sensitivity to Guidance Strength}
\label{app:guidance_strength}

\begin{figure}[htbp]
    \centering
    \includegraphics[
        width=0.42\textwidth
    ]{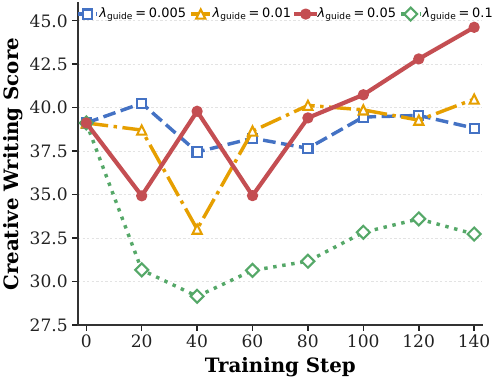}
    \caption{Post-hoc sensitivity analysis of the guidance-strength coefficient $\lambda_{\mathrm{guide}}$ on CreativeWriting-V3 using Qwen3-4B. Each variant is trained for 140 steps; checkpoints saved every 20 steps are subsequently evaluated. The setting $\lambda_{\mathrm{guide}}=0.05$, selected based on training-reward improvement, also achieves the highest final benchmark score among the tested values.}
    \label{fig:guidance_strength}
\end{figure}



We select $\lambda_{\mathrm{guide}}$ using 140-step preliminary training runs with Qwen3-4B in the Writing domain, comparing $\lambda_{\mathrm{guide}}\in\{0.005,0.01,0.05,0.1\}$. The selection criterion is the increase in training reward over these runs. Among the tested values, $\lambda_{\mathrm{guide}}=0.05$ yields the largest training-reward increase and is therefore fixed for the full Writing training runs with both Qwen3-4B and Qwen3-14B before the downstream sensitivity evaluation.

For Health and Science, we apply the same 140-step preliminary selection procedure on Qwen3-4B over the same candidate set, using training-reward improvement as the selection criterion. This procedure selects $\lambda_{\mathrm{guide}}=0.01$ for both domains, which is then fixed for the full training runs at both model scales. As in Writing, no downstream benchmark scores are used for coefficient selection.

After fixing the coefficient, we evaluate checkpoints saved every 20 steps from these preliminary runs on CreativeWriting-V3 as a post-hoc sensitivity analysis. As shown in Fig.~\ref{fig:guidance_strength}, $\lambda_{\mathrm{guide}}=0.05$ also achieves the highest final CreativeWriting-V3 score among the tested values. A small coefficient ($0.005$) leaves performance near the initial baseline, indicating that the guidance signal is too weak to meaningfully affect policy optimization. In contrast, a large coefficient ($0.1$) causes a sharp early performance drop; although the model partially recovers, it remains below the moderate setting, suggesting that overly strong guidance disrupts the existing policy distribution and makes subsequent recovery difficult. These downstream results are consistent with the training-reward-based selection: $\lambda_{\mathrm{guide}}=0.05$ yields both the largest training-reward increase and the highest final CreativeWriting-V3 score among the tested values, providing additional support for the selected coefficient.

CreativeWriting-V3 scores are not used to select or readjust the coefficient. We do not perform a separate coefficient sweep on Qwen3-14B or readjust the coefficient based on the full-run benchmark results.

\subsection{Sensitivity to Guidance Removal Timing}
\label{app:guidance_removal_timing}

We study the sensitivity of \textsc{RISE-RL} to the guidance
removal step using Qwen3-14B in the Health domain.
In addition to the default removal step
$s_{\mathrm{switch}}=60$, we evaluate removal at steps
20, 40, and 100.
We include two boundary conditions:
$s_{\mathrm{switch}}=0$, corresponding to standard Rubric-RL
without guidance, and $s_{\mathrm{switch}}=\infty$,
corresponding to persistent guidance throughout training.
We report HealthBench, LLMEval-Med, and MedQA scores,
together with their unweighted mean.

\paragraph{Sensitivity around the Default Setting.}
As shown in Table~\ref{tab:guidance_removal_timing},
removing guidance at steps 40 and 100 yields average scores
of 72.39 and 71.99, respectively, compared with 73.11
at the default step of 60.
Both alternatives retain an average improvement over
Rubric-RL, whose score is 70.85.
On HealthBench, the scores at steps 40 and 100 are
particularly close to the default, differing by only
0.18 and 0.29 points, respectively.
The differences are larger on the other benchmarks,
indicating that sensitivity varies across tasks.
In this setting, all tested finite removal steps from
20 to 100 outperform Rubric-RL in average performance,
while steps 40--100 provide consistently strong results.
This broad effective range shows that the improvement is
not tied to a narrowly tuned removal checkpoint.

\paragraph{Early Removal and Persistent Guidance.}
Removing guidance at step 20 yields an average score
of 72.00, improving over Rubric-RL but falling below
the default setting.
This is consistent with early removal limiting the benefit
of guided training, although its average performance
is comparable to removal at step 100.
Persistent guidance yields an average score of 71.06
and reduces HealthBench performance to 55.96,
below the Rubric-RL score of 57.92.
Thus, retaining guidance throughout training does not
provide the same benefit as removing it at an intermediate
stage.
Together, these comparisons support an initial guidance
phase followed by unguided optimization.

\begin{table}[t]
    \centering
    \caption{
        Sensitivity to guidance removal timing on Qwen3-14B
        in Health.
        A removal step of $0$ denotes Rubric-RL without
        guidance, and $\infty$ denotes persistent guidance.
        Avg.\ is the unweighted mean of the three benchmark
        scores. Best scores are bold.
    }
    \label{tab:guidance_removal_timing}
    \small
    \renewcommand{\arraystretch}{1.15}
    \setlength{\tabcolsep}{5pt}

    \begin{tabular*}{\linewidth}{
        @{\extracolsep{\fill}} lcccc @{}
    }
        \toprule
        Removal step
        & HealthBench
        & LLMEval-Med
        & MedQA
        & Avg. \\
        \midrule
        $0$ (Rubric-RL)
        & 57.92 & 76.40 & 78.22 & 70.85 \\
        \midrule
        $20$
        & 58.51 & 76.82 & 80.67 & 72.00 \\
        $40$
        & 59.26 & 77.54 & 80.38 & 72.39 \\
        $60$ (default)
        & \textbf{59.44}
        & \textbf{78.31}
        & \textbf{81.57}
        & \textbf{73.11} \\
        $100$
        & 59.15 & 76.22 & 80.59 & 71.99 \\
        \midrule
        $\infty$ (persistent)
        & 55.96 & 76.98 & 80.23 & 71.06 \\
        \bottomrule
    \end{tabular*}
\end{table}

\subsection{Automatic Guidance Removal}
\label{app:automatic_guidance_removal}

Different models and domains may exhibit different learning rates,
making a fixed removal step difficult to transfer across settings.
We therefore formulate guidance removal in terms of observed training
progress rather than the absolute optimization step. Since rubric
rewards are normalized to $[0,1]$ across domains, we remove guidance
when the training reward exceeds its initial value by $0.1$ and
continue training with natural-prompt GRPO. The shared normalized
reward scale makes this criterion directly comparable across domains
without selecting a separate removal step for each setting.

We choose $0.1$ as a practical transition point based on the observed
training dynamics. A substantially smaller increment may remove
guidance before the targeted behaviors have been sufficiently
internalized, limiting the benefit of selective exploration. In
contrast, requiring a much larger increment prolongs the more
expensive guidance stage after its marginal benefit has begun to
diminish and leaves less opportunity for subsequent autonomous
optimization. This trade-off is consistent with the removal-timing
and persistent-guidance results in
Appendix~\ref{app:guidance_removal_timing}.

As shown in Table~\ref{tab:automatic_guidance_removal}, the same
reward-progress threshold, applied without domain-specific adjustment,
matches manual scheduling in Writing and outperforms it in Science.
Together with the broad effective range observed in the Health timing
analysis, these results indicate that the gains of RISE-RL are not
tied to a narrowly selected manual checkpoint. They further support
normalized reward progress as a simple, transferable, and reproducible
mechanism for transitioning from guided to autonomous optimization.

\begin{table}[t]
    \centering
    \caption{
        Automatic versus manual guidance removal on
        Qwen3-4B.
        Automatic removal uses a shared threshold of the initial normalized reward plus $0.1$, without domain-specific adjustment.
        Avg.\ denotes the unweighted mean of the two
        benchmarks within each domain.
        Best scores within each domain are bold.
    }
    \label{tab:automatic_guidance_removal}
    \small
    \renewcommand{\arraystretch}{1.15}
    \setlength{\tabcolsep}{6pt}

    \begin{tabular*}{\linewidth}{
        @{\extracolsep{\fill}} lccc @{}
    }
        \toprule
        \multicolumn{4}{c}{Writing} \\
        \cmidrule(lr){1-4}
        Removal strategy
        & WritingBench
        & CreativeWriting-V3
        & Avg. \\
        \midrule
        Manual
        & \textbf{72.15}
        & 46.73
        & 59.44 \\
        Automatic
        & 71.97
        & \textbf{47.02}
        & \textbf{59.50} \\
        \midrule
        \multicolumn{4}{c}{Science} \\
        \cmidrule(lr){1-4}
        Removal strategy
        & ResearchQA
        & GPQA-Diamond
        & Avg. \\
        \midrule
        Manual
        & \textbf{77.08}
        & 49.12
        & 63.10 \\
        Automatic
        & 76.54
        & \textbf{52.53}
        & \textbf{64.54} \\
        \bottomrule
    \end{tabular*}
\end{table}

\subsection{Training Dynamics of RuscaRL}
\label{subsec:appendix-ruscal}


\begin{figure}
    \centering
    \includegraphics[width=\linewidth]{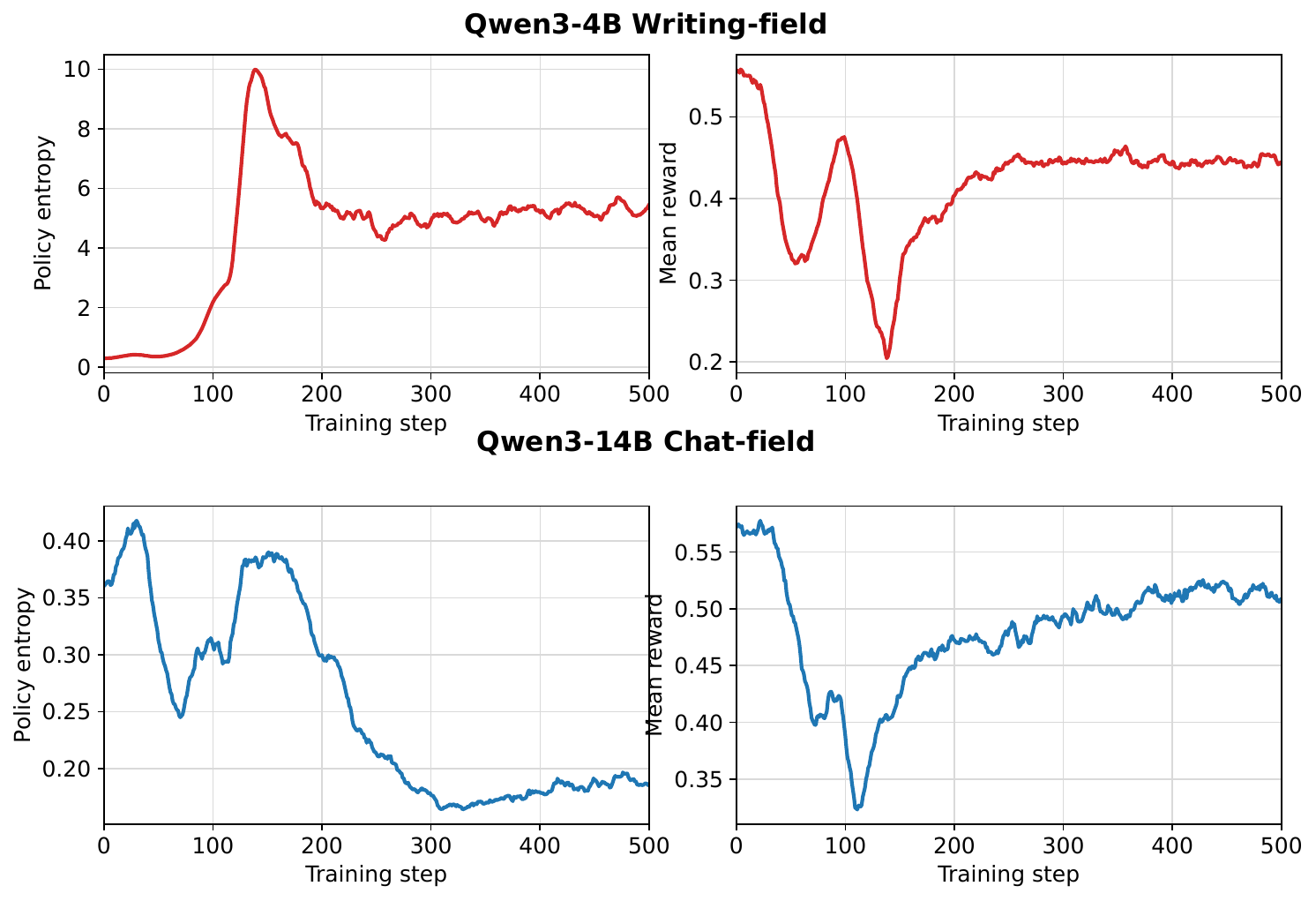}
    \caption{Representative training dynamics of RuscaRL under two model--domain settings: Qwen3-4B in the Writing domain (top) and Qwen3-14B in the Chat domain (bottom). The left column shows policy entropy, while the right column shows the mean rubric reward over 500 training steps.}
    \label{fig:ruscarl_dynamics}
\end{figure}

We further examine the training dynamics of RuscaRL under two
representative model--domain settings, as shown in
Figure~\ref{fig:ruscarl_dynamics}. The corresponding reproduction
configuration is provided in Section~\ref{app:training_configuration}.
During the initial stage, both settings maintain relatively high mean
rewards with rubric scaffolding. However, their entropy dynamics differ:
Qwen3-4B trained on Writing exhibits a sharp increase in policy entropy,
whereas Qwen3-14B trained on Chat fluctuates around its initial entropy
level. When the scaffolding is substantially removed after approximately
100 training steps, both settings experience an abrupt reward collapse.
Although the reward gradually recovers during the subsequent unscaffolded
training stage, it remains below its initial scaffolded level by the end
of training.

One possible interpretation is that removing scaffolding induces a mismatch between scaffold-conditioned and natural-prompt generation. This transition may contribute to the observed reward drop. In the early stage, extensive rubric
guidance makes high-reward responses easier to generate. Nevertheless,
injecting many criteria into the prompt may dilute adherence to the original
user instruction and induce trajectories that differ substantially from
those sampled under the natural prompt. Jointly optimizing these
heterogeneous trajectories within the same group-relative objective may
therefore produce unstable policy updates, as reflected by the abrupt or
fluctuating entropy dynamics. Once the scaffolding is removed, the policy
must transition from rubric-conditioned generation to natural-prompt
generation, causing a pronounced distribution shift and the accompanying
reward collapse. Later on-policy training can partially recover performance,
but the instability introduced during this transition may constrain the
final performance attainable by the model.

\section{Theoretical Analysis}
\label{app:theoretical_analysis}

\subsection{A Hybrid View of SFT and On-Policy Distillation}
\label{app:hybrid_sft_opd}
This section provides an alternative interpretation of our selective
guidance objective. We first review supervised fine-tuning (SFT) and
on-policy distillation (OPD), and then show that our guidance branch
combines the online, policy-dependent supervision of OPD with the
hard-target optimization structure of SFT.

\paragraph{Supervised fine-tuning.}

Let $q$ denote an input prompt and
$o=(o_1,\ldots,o_{T(o)})$ denote a reference response, where $T(o)$
is its token length. In practice, standard SFT commonly minimizes the
length-normalized token-level negative log-likelihood
\begin{equation}
\mathcal{L}_{\mathrm{SFT}}(\theta)
=
-\mathbb{E}_{(q,o)\sim\mathcal{D}_{\mathrm{SFT}}}
\left[
\frac{1}{T(o)}
\sum_{t=1}^{T(o)}
\log
\pi_\theta(o_t\mid q,o_{<t})
\right].
\label{eq:sft_objective}
\end{equation}

At each prefix $(q,o_{<t})$, the reference token $o_t$ defines an
empirical one-hot target distribution
\begin{equation}
\delta_{o_t}(v)
=
\mathbb{I}[v=o_t].
\end{equation}
The corresponding forward KL divergence is
\begin{align}
D_{\mathrm{KL}}
\left(
\delta_{o_t}
\Vert
\pi_\theta(\cdot\mid q,o_{<t})
\right)
&=
\sum_{v}
\delta_{o_t}(v)
\log
\frac{\delta_{o_t}(v)}
{\pi_\theta(v\mid q,o_{<t})}
\nonumber\\
&=
-\log
\pi_\theta(o_t\mid q,o_{<t}),
\label{eq:sft_forward_kl}
\end{align}
where the equality is exact because the entropy of the one-hot target
distribution is zero. Thus, SFT can be viewed as forward-KL
distillation from an empirical one-hot target distribution.

More generally, if reference responses are sampled from a fixed
teacher or behavior distribution $\mu$, then
\begin{equation}
\mathbb{E}_{o\sim\mu(\cdot\mid q)}
[-\log\pi_\theta(o\mid q)]
=
H(\mu(\cdot\mid q))
+
D_{\mathrm{KL}}
\left(
\mu(\cdot\mid q)
\Vert
\pi_\theta(\cdot\mid q)
\right).
\label{eq:sft_distribution_view}
\end{equation}
Eq.~(\ref{eq:sft_distribution_view}) is a sequence-level identity,
whereas Eq.~(\ref{eq:sft_objective}) averages the token losses by
$T(o)$. The latter is therefore a practical length-normalized empirical
objective rather than an exact sequence-level KL identity when response
lengths vary.

\paragraph{On-policy distillation.}

In standard OPD, trajectories are sampled from the current student
policy, while a teacher provides dense token-level distributional
supervision on the student-visited prefixes. Let
$\pi_T$ denote the teacher policy. A sequence-level reverse-KL
objective can be written as
\begin{equation}
\mathcal{L}_{\mathrm{OPD}}(\theta)
=
\mathbb{E}_{q\sim\mathcal{D}}
\left[
D_{\mathrm{KL}}
\left(
\pi_\theta(\cdot\mid q)
\Vert
\pi_T(\cdot\mid q)
\right)
\right].
\label{eq:opd_sequence}
\end{equation}

By the autoregressive chain rule for KL divergence, the sequence-level
objective factorizes exactly as
\[
\mathcal{L}_{\mathrm{OPD}}(\theta)
=
\mathbb{E}_{
q\sim\mathcal{D},
\,o\sim\pi_\theta(\cdot\mid q)
}
\left[
\sum_{t=1}^{T(o)}
D_{\mathrm{KL}}
\left(
\pi_\theta(\cdot\mid q,o_{<t})
\Vert
\pi_T(\cdot\mid q,o_{<t})
\right)
\right].
\]

In practical token-level optimization, we use the corresponding
length-normalized surrogate
\begin{equation}
\widetilde{\mathcal{L}}_{\mathrm{OPD}}(\theta)
=
\mathbb{E}_{
q\sim\mathcal{D},
\,o\sim\pi_\theta(\cdot\mid q)
}
\left[
\frac{1}{T(o)}
\sum_{t=1}^{T(o)}
D_{\mathrm{KL}}
\left(
\pi_\theta(\cdot\mid q,o_{<t})
\Vert
\pi_T(\cdot\mid q,o_{<t})
\right)
\right].
\label{eq:opd_token}
\end{equation}
Eq.~(\ref{eq:opd_token}) stabilizes the contribution of responses
with different lengths, but it is not algebraically identical to the
unnormalized sequence-level reverse KL in
Eq.~(\ref{eq:opd_sequence}).

OPD therefore differs from conventional SFT in two respects. First,
its state distribution is induced by the current student policy rather
than by a fixed offline dataset. Second, it typically matches a soft
teacher distribution through reverse KL instead of treating a single
reference token as a hard target.

\paragraph{Online construction of privileged reference trajectories.}

Our guidance branch begins with online, policy-dependent trajectory
generation. Given the selected feedback criteria
$\mathcal{C}^{\mathrm{fb}}(q)$, we construct
\begin{equation}
q^{\mathrm{priv}}
=
q\oplus\mathcal{C}^{\mathrm{fb}}(q)
\end{equation}
and sample privileged trajectories from the current behavior policy:
\begin{equation}
o_j^{\mathrm{priv}}
\sim
\pi_{\theta_{\mathrm{old}}}
(\cdot\mid q^{\mathrm{priv}}).
\label{eq:privileged_rollout_appendix}
\end{equation}

Unlike an offline SFT dataset, these trajectories are regenerated as
the policy evolves. We evaluate each privileged trajectory using the
complete original rubric. Let $\bar r^{\mathrm{nat}}$ and
$\sigma^{\mathrm{nat}}$ denote the mean and standard deviation of the
natural-rollout rewards $\{r_i^{\mathrm{nat}}\}_{i=1}^{G}$,
respectively, and let $\epsilon_A=10^{-6}$. We define the normalized
reward gain and retain trajectories with positive gains:
\begin{equation}
A_j^{\mathrm{ref}}
=
\frac{r_j^{\mathrm{priv}}-\bar r^{\mathrm{nat}}}
{\sigma^{\mathrm{nat}}+\epsilon_A},
\qquad
\mathcal{O}^{\mathrm{ref}}
=
\left\{
o_j^{\mathrm{priv}}
\mid
A_j^{\mathrm{ref}}>0
\right\}.
\label{eq:quality_filter_appendix}
\end{equation}

The resulting reference set is therefore both policy-dependent and
quality-filtered. This online construction is related to OPD in that
the supervision distribution evolves together with the current
policy. However, the trajectories are generated under privileged
conditioning rather than from the natural-prompt student policy.

\paragraph{Support-adaptive SFT interpretation.}

After removing the privileged criteria, each retained trajectory is
teacher-forced under the original prompt. For a retained token
$o_{j,t}^{\mathrm{ref}}$, define its natural-policy support as
\begin{equation}
p_{j,t}
=
\pi_\theta
\left(
o_{j,t}^{\mathrm{ref}}
\mid
q,o_{j,<t}^{\mathrm{ref}}
\right).
\label{eq:natural_support_appendix}
\end{equation}

Our policy-support transformation is
\begin{equation}
\rho_{j,t}
=
\frac{p_{j,t}}{p_{j,t}+\gamma},
\label{eq:support_transform_appendix}
\end{equation}
and the corresponding token-level loss is
\begin{equation}
\ell_{\mathrm{guide}}(p_{j,t})
=
-\log\rho_{j,t}
=
-\log p_{j,t}
+
\log(p_{j,t}+\gamma).
\label{eq:token_guidance_appendix}
\end{equation}

The following result establishes its connection to SFT.

\paragraph{Proposition 1.}
For a fixed reference token, the gradient of the support-shaped
guidance loss has the same direction as the gradient of the standard
SFT negative log-likelihood:
\begin{equation}
\nabla_\theta
\ell_{\mathrm{guide}}(p_{j,t})
=
\frac{\gamma}{p_{j,t}+\gamma}
\nabla_\theta
\left[-\log p_{j,t}\right].
\label{eq:gradient_equivalence}
\end{equation}

\paragraph{Proof.}
Starting from Eq.~(\ref{eq:token_guidance_appendix}),
\begin{align}
\nabla_\theta\ell_{\mathrm{guide}}(p_{j,t})
&=
\left(
-\frac{1}{p_{j,t}}
+
\frac{1}{p_{j,t}+\gamma}
\right)
\nabla_\theta p_{j,t}
\nonumber\\
&=
-\frac{\gamma}
{p_{j,t}(p_{j,t}+\gamma)}
\nabla_\theta p_{j,t}
\nonumber\\
&=
\frac{\gamma}{p_{j,t}+\gamma}
\nabla_\theta[-\log p_{j,t}].
\end{align}
\hfill$\square$

Eq.~(\ref{eq:gradient_equivalence}) shows that the guidance loss is
gradient-equivalent to an SFT update with the adaptive token weight
\begin{equation}
w_{j,t}^{\mathrm{support}}
=
\frac{\gamma}{p_{j,t}+\gamma}.
\label{eq:support_gradient_weight}
\end{equation}
Importantly, this is a gradient-level equivalence; the guidance loss
itself is not equal to a weighted SFT loss unless the adaptive weight
is treated as a stop-gradient quantity.

For weakly supported tokens, $p_{j,t}\ll\gamma$, we have
\begin{equation}
w_{j,t}^{\mathrm{support}}
\approx 1,
\qquad
\ell_{\mathrm{guide}}(p_{j,t})
\approx
-\log p_{j,t}+\log\gamma.
\end{equation}
The resulting update therefore approaches a standard SFT update. In
contrast, when $p_{j,t}\gg\gamma$,
\begin{equation}
w_{j,t}^{\mathrm{support}}
\approx
\frac{\gamma}{p_{j,t}},
\end{equation}
which suppresses imitation of tokens already well supported by the
natural policy.

For each retained trajectory
$o_j^{\mathrm{ref}}\in\mathcal{O}^{\mathrm{ref}}$,
let $T_j$ denote its token length. We define the trajectory-level
weight as
$\widetilde A_j=\min\!\left(A_j^{\mathrm{ref}},A_{\max}\right)$,
where $A_{\max}=5$ limits large normalized gains when the
natural-rollout reward variance is small.
The selective guidance loss is then
\begin{equation}
\mathcal{L}_{\mathrm{guide}}
=
-
\frac{1}{|\mathcal O^{\mathrm{ref}}|}
\sum_{o_j^{\mathrm{ref}}\in\mathcal O^{\mathrm{ref}}}
\frac{\widetilde A_j}{T_j}
\sum_{t=1}^{T_j}
\log \rho_{j,t}.
\label{eq:selective_guidance_loss_appendix}
\end{equation}

Its gradient can consequently be interpreted as
\begin{equation}
\nabla_\theta\mathcal{L}_{\mathrm{guide}}
=
\frac{1}{|\mathcal{O}^{\mathrm{ref}}|}
\sum_{o_j^{\mathrm{ref}}\in\mathcal{O}^{\mathrm{ref}}}
\frac{\widetilde A_j}{T_j}
\sum_{t=1}^{T_j}
w_{j,t}^{\mathrm{support}}
\nabla_\theta[-\log p_{j,t}].
\label{eq:weighted_sft_gradient}
\end{equation}

Thus, the guidance branch admits a gradient-level interpretation as trajectory-quality-weighted, token-support-adaptive negative-log-likelihood training on online privileged trajectories. This interpretation concerns only the auxiliary branch; the full training procedure additionally retains natural-prompt GRPO optimization.

\paragraph{Connection and distinction.}

Our guidance branch combines complementary aspects of SFT and OPD.
Similar to OPD, the supervision is online and policy-dependent: the
privileged reference trajectories are regenerated as the behavior
policy evolves. Similar to SFT, the optimization treats the realized
tokens in the retained trajectories as hard targets and follows the
token-level negative-log-likelihood direction.

Nevertheless, our method is not equivalent to conventional OPD.
Standard OPD samples natural-prompt trajectories from the student and
matches the full teacher distribution, typically through reverse KL.
Our method instead samples trajectories under privileged rubric
conditioning, filters them using the complete-rubric reward, removes
the privileged information, and internalizes their realized tokens
through a support-adaptive hard-target objective. It therefore does
not require access to full teacher logits.

In summary, our guidance branch combines 
(i) online, policy-dependent reference construction, which is OPD-like;
(ii) hard-target teacher-forcing optimization, which is SFT-like; and
(iii) quality filtering and policy-support shaping, which enable the
selective internalization of privileged trajectories.

\section{Prompt and Rubric Templates}
\label{app:prompt_templates}



\subsection{Selective Guidance Prompt}
\label{app:guidance_prompt}


During guidance, RISE-RL preserves the original conversation and
does not replace or rewrite the user prompt. Instead, it constructs a
lightweight privileged suffix from the rubric evaluation of the initial
natural-rollout group. This suffix provides targeted guidance for criteria that the
current policy fails to satisfy consistently.

For each criterion $c_k$, we compute the priority score
\[
    p_k = w_k f_k,
\]
where $f_k$ is the number of natural-rollout responses that fail
criterion $c_k$. We then select
\[
    C^{\mathrm{fb}}(q)
    =
    \operatorname{TopM}
    \left(C(q);\{p_k\}_{k=1}^{K}\right),
    \qquad M=5.
\]
Only criteria with $p_k>0$ are retained; if no such criterion exists, the guidance branch is skipped.

The textual descriptions of the selected criteria are concatenated
using semicolons to form the criterion-level hint. One of five natural
language templates is then sampled at random, filled with this hint,
and appended to the end of the final user message in the original
conversation. Thus, the re-rollout prompt preserves the original task
while introducing only a small amount of failure-dependent privileged
information.

\paragraph{Guidance Templates.}

For every re-rollout, one of the following five templates is sampled
at random. The placeholder \texttt{\{hint\}} is replaced by the
descriptions of at most five selected rubric criteria.

\begin{tcolorbox}[
    breakable,
    colback=risebluebg,
    colframe=riseblue,
    boxrule=0.7pt,
    left=6pt,
    right=6pt,
    top=5pt,
    bottom=5pt
]
\small
\begin{enumerate}
    \item
    \textit{When answering, please pay attention to the following
    aspects: \texttt{\{hint\}}}

    \item
    \textit{For reference, the following aspects deserve particular
    attention: \texttt{\{hint\}}}

    \item
    \textit{You may consider addressing the question from the following
    perspectives: \texttt{\{hint\}}}

    \item
    \textit{The following points may help improve the quality of the
    response: \texttt{\{hint\}}}

    \item
    \textit{Please appropriately reflect the following points in your
    response: \texttt{\{hint\}}}
\end{enumerate}
\end{tcolorbox}

\paragraph{Prompt Composition.}

The original user message is preserved verbatim. For each guided
re-rollout, RISE-RL appends one randomly sampled instruction template
whose placeholder is filled with at most five high-priority failed
criteria.

Importantly, the privileged suffix contains only a small subset of the
instance-specific rubric rather than the complete rubric. Its content
is constructed adaptively from the criterion-level failure pattern
observed across the initial natural rollouts, and it is used only to generate the re-rollout trajectories.
The natural-prompt trajectories remain unchanged. Consequently, the
additional information serves as targeted exploratory guidance rather
than a replacement for the original task.

This construction differs from static rubric scaffolding in two
important respects. First, the guidance is selective: only the most
persistent and high-value failed criteria are included. Second, the
guidance is adaptive: the selected criteria depend on the behavior of
the current policy on the specific prompt. This enables RISE-RL to
direct exploration toward behaviors that are both important and
unlikely to be discovered through unguided sampling alone.

\subsection{Diversity Evaluation Prompt}
\label{app:diversity-judge-prompt}

The following template is used for pairwise diversity evaluation.
The placeholders \texttt{\{prompt\}},
\texttt{\{a1\}}--\texttt{\{a6\}}, and
\texttt{\{b1\}}--\texttt{\{b6\}} are replaced by the
writing prompt and the responses in their randomized
presentation order.

\begin{tcolorbox}[
    breakable,
    colback=risegraybg,
    colframe=risegray,
    title={\textbf{Diversity Evaluation Prompt}},
    fonttitle=\small,
    boxrule=0.7pt,
    left=5pt,
    right=5pt,
    top=4pt,
    bottom=4pt
]
\small
\setlength{\parindent}{0pt}

You are comparing EXPRESSION-LEVEL DIVERSITY between two
anonymous sets of creative-writing responses.

\medskip
You will receive one creative-writing prompt and two
response sets, Set A and Set B.
Each set contains six independently generated responses
to the same prompt.

\medskip
Your task is comparative: determine which set expresses
the prompt-required content in more substantively
different ways.
Do not evaluate which set has the best individual response.

\medskip
\textbf{IMPORTANT EVALUATION TARGET}

The prompt may already constrain the central premise,
characters, events, themes, genre, or intended outcome.
Do NOT treat such prompt-imposed similarities as evidence
of low diversity.
Do NOT require the responses to adopt different top-level
plots or themes.

\medskip
Instead, evaluate whether the responses realize the same
required content through meaningfully different:
\begin{itemize}
    \setlength{\itemsep}{0pt}
    \setlength{\parsep}{0pt}
    \item linguistic formulations,
    \item discourse organizations,
    \item perspectives and focalization strategies,
    \item tones and narrative voices,
    \item imagery and detail presentation.
\end{itemize}

\medskip
\textbf{GENERAL PRINCIPLES}

\begin{enumerate}
    \setlength{\itemsep}{2pt}
    \setlength{\parsep}{0pt}

    \item Reward substantive variation in expression and
    presentation, not deviation from the prompt.

    \item Do not reward superficial substitutions in names,
    locations, isolated objects, synonyms, or minor wording.

    \item Do not reward incoherent, irrelevant, incomplete,
    or instruction-violating responses as diversity.

    \item Judge diversity independently of average writing
    quality, except that invalid responses should not
    contribute positively.

    \item Compare the two sets directly.
    Use the full preference scale when the evidence
    supports it; do not default to ties.

    \item Ignore the presentation order of the sets
    and responses.
\end{enumerate}

\medskip
\textbf{DIVERSITY DIMENSIONS}

\medskip
\textbf{A.}
\texttt{linguistic\_realization\_diversity}

Differences in lexical choice, sentence construction,
syntactic patterns, rhetorical devices, rhythm,
compression versus elaboration, directness versus
implication, and figurative language.
Simple synonym substitution does not count unless it
changes the overall expression strategy.

\medskip
\textbf{B.}
\texttt{discourse\_organization\_diversity}

Differences in openings, paragraph structure, sequencing
of information or events, placement of exposition and
turning points, transitions, pacing, scene organization,
and closing strategies.

\medskip
\textbf{C.}
\texttt{perspective\_focalization\_diversity}

Differences in narrative viewpoint, focal character,
psychological versus external focus, information access,
distance from events, and timing of information disclosure.
A superficial switch of pronouns does not count unless
it changes the reader's experience.

\medskip
\textbf{D.}
\texttt{tone\_voice\_diversity}

Differences in emotional tone, narrative voice,
stylistic stance, formality, humor, irony, restraint,
intensity, and emotional distance.

\medskip
\textbf{E.}
\texttt{imagery\_detail\_realization\_diversity}

Differences in sensory imagery, scene construction,
dialogue use, action detail, symbolic objects,
descriptive focus, and how the responses show rather
than merely state ideas.

\medskip
\textbf{PAIRWISE MARGIN SCALE}

For each dimension, return one integer margin:

\smallskip
$+3$ = Set A is substantially more diverse\par
$+2$ = Set A is clearly more diverse\par
$+1$ = Set A is slightly more diverse\par
$\phantom{+}0$ = no meaningful difference\par
$-1$ = Set B is slightly more diverse\par
$-2$ = Set B is clearly more diverse\par
$-3$ = Set B is substantially more diverse\par

\medskip
Use 0 only when the two sets are genuinely
indistinguishable on that dimension.
A difference in only one response generally supports
at most a margin of 1, unless that response introduces
a clearly distinct expression strategy absent from
the other set.

\medskip
Also return:
\begin{itemize}
    \setlength{\itemsep}{0pt}
    \setlength{\parsep}{0pt}

    \item \texttt{overall\_margin}: holistic comparison
    on the same -3 to +3 scale.

    \item \texttt{confidence}: a number from 0.0 to 1.0.

    \item \texttt{invalid\_response\_ids} for each set.

    \item concise evidence describing the main
    comparative differences.
\end{itemize}

\medskip
Return JSON only:

\begin{verbatim}
{
  "set_A_invalid_response_ids": [],
  "set_B_invalid_response_ids": [],
  "linguistic_realization_diversity": {
    "margin": 0, "evidence": ""
  },
  "discourse_organization_diversity": {
    "margin": 0, "evidence": ""
  },
  "perspective_focalization_diversity": {
    "margin": 0, "evidence": ""
  },
  "tone_voice_diversity": {
    "margin": 0, "evidence": ""
  },
  "imagery_detail_realization_diversity": {
    "margin": 0, "evidence": ""
  },
  "overall_margin": 0,
  "confidence": 0.0,
  "comparative_summary": ""
}
\end{verbatim}

\medskip
\textbf{CREATIVE-WRITING PROMPT:}

\texttt{\{prompt\}}

\medskip
\textbf{SET A:}

\smallskip
[A1]\par
\texttt{\{a1\}}

\smallskip
[A2]\par
\texttt{\{a2\}}

\smallskip
[A3]\par
\texttt{\{a3\}}

\smallskip
[A4]\par
\texttt{\{a4\}}

\smallskip
[A5]\par
\texttt{\{a5\}}

\smallskip
[A6]\par
\texttt{\{a6\}}

\medskip
\textbf{SET B:}

\smallskip
[B1]\par
\texttt{\{b1\}}

\smallskip
[B2]\par
\texttt{\{b2\}}

\smallskip
[B3]\par
\texttt{\{b3\}}

\smallskip
[B4]\par
\texttt{\{b4\}}

\smallskip
[B5]\par
\texttt{\{b5\}}

\smallskip
[B6]\par
\texttt{\{b6\}}

\end{tcolorbox}

\section{Case Studies}
\label{app:case_studies}

\subsection{Creative-Writing Case Study}
\label{app:writing_case}
\label{app:cw-evaluation}

\paragraph{Evaluation setup.}
We evaluate all 32 prompts in CreativeWriting-V3, with
six independently generated responses per method and
prompt. Writing quality is assessed using Claude Sonnet 4.
For diversity, we compare the initial model, Rubric-RL,
and RISE-RL using DeepSeek-V4-Pro-Low-Latency with
temperature set to 0. For each prompt, every pair of
methods is compared by presenting their six-response
sets anonymously to the judge. Each response is capped
at its first 6,000 characters, followed by a truncation
marker when applicable.

\paragraph{Diversity dimensions.}
The judge assesses expression-level diversity along five
dimensions:
\begin{itemize}
    \setlength{\itemsep}{2pt}
    \setlength{\parsep}{0pt}

    \item \textbf{Linguistic realization:}
    variation in lexical choice, sentence construction,
    syntax, rhetorical devices, and rhythm.

    \item \textbf{Discourse organization:}
    variation in openings, paragraph structure,
    information sequencing, pacing, and closing strategies.

    \item \textbf{Perspective and focalization:}
    variation in narrative viewpoint, focal character,
    psychological versus external focus, and information
    disclosure.

    \item \textbf{Tone and voice:}
    variation in emotional tone, narrative voice,
    formality, humor, irony, and emotional distance.

    \item \textbf{Imagery and detail realization:}
    variation in sensory imagery, scene construction,
    dialogue, action details, and descriptive focus.
\end{itemize}

These dimensions capture substantive variation in how
responses fulfill the same prompt, without requiring
different plots or themes when these are constrained
by the prompt. The judge is instructed not to reward
superficial substitutions, incoherence, irrelevance,
incompleteness, or instruction violations.
The full judging prompt and dimension definitions are
provided in Appendix~\ref{app:diversity-judge-prompt}.

\paragraph{Pairwise scoring.}
Each method pair is evaluated under three randomized
presentation orders (seeds 0, 1, and 2), varying both
the order of the two sets and the response order within
each set. For each dimension, the judge returns an
integer preference margin from $-3$ to $+3$:
positive values favor Set A, negative values favor
Set B, and 0 indicates no meaningful difference.
Absolute margins of 1, 2, and 3 indicate slight, clear,
and substantial differences, respectively.
The judge also returns a separate holistic margin.
The overall diversity score uses this holistic judgment
rather than an average of the five dimension-specific
judgments.

\paragraph{Score aggregation and uncertainty.}
After restoring method orientation so that a positive
margin favors method $m$ over method $n$, we average
the valid margins across presentation orders to obtain
$\bar{\delta}_{q,mn}$ for prompt $q$.
The normalized pairwise preference scores are
\begin{equation}
    s_{q,mn}
    =
    \frac{1}{2}
    +
    \frac{\bar{\delta}_{q,mn}}{6},
    \qquad
    s_{q,nm}
    =
    1-s_{q,mn}.
    \label{eq:cw-diversity-preference}
\end{equation}
These scores lie in $[0,1]$.
For each method, we first average over available
opponents within each prompt, then equally over
prompts with valid scores, yielding the aggregate
score $S_m$. The same aggregation is applied separately
to the overall margin and each dimension-specific margin.

The evaluation script computes 95\% percentile
confidence intervals for these aggregate scores using
2,000 prompt-level bootstrap resamples
(bootstrap seed 12345).
The resampling unit is the prompt, with the
presentation-order judgments aggregated before
resampling.

\paragraph{Relative gains in Figure~4.}
For the diversity results in Figure~4, we further
convert aggregate normalized preference scores into
relative gains over the initial Qwen3-4B model:
\begin{equation}
    G_m
    =
    \frac{S_m-S_{\mathrm{base}}}
         {S_{\mathrm{base}}}
    \times 100\%.
    \label{eq:cw-diversity-relative-gain}
\end{equation}
Here, $S_{\mathrm{base}}$ is the initial model's
aggregate overall diversity preference score under
the same comparison protocol and comparison set.
The reported diversity gain therefore represents
a relative increase in this normalized preference score.

\paragraph{Qualitative case study.}
We conduct a qualitative comparison on a representative creative-writing
prompt that asks the model to portray a non-combat slice of a Roman
gladiator's daily life in first-person, past-tense narration, while
incorporating sensory detail, internal emotion, and the political and social
context of the Roman Empire. The selected modifier further requires the
response to describe the gladiator's preferred weapon and explain its
personal significance. We present one complete response from the Initial, SFT,
Rubric-RL, and RISE-RL models. Green highlights mark representative strengths, while red highlights mark representative failure modes discussed below.

\paragraph{Qualitative comparison.}
On this representative example, RISE-RL achieves the strongest overall
performance across the four models. It obtains the best score on 12 of
the 13 positive criteria, with particularly clear advantages in
instruction adherence, character plausibility, imagery, coherence,
emotional engagement, and overall impression. These gains indicate
that RISE-RL improves not only surface-level style, but also the
narrative structure, thematic consistency, and emotional depth of the
response.

RISE-RL also performs strongly on the negative criteria, achieving the
lowest score on Meandering, Amateurish, Purple Prose, and Overwrought,
while tying for the best result on Tell-Don't-Show. Taken together, the
results show that RISE-RL produces a more vivid, coherent, and engaging
response while simultaneously reducing several common failure modes in
creative writing. Although a small number of dimensions exhibit
sample-level variation, the overall pattern consistently favors
RISE-RL. The response is nevertheless not flawless: it still includes
a combat sequence despite the non-combat instruction and appends a
brief meta-level summary after the story. These localized deviations
coexist with broader gains in imagery, characterization, thematic
integration, and overall reader engagement. The corresponding
criterion-level results are reported in
Table~\ref{tab:case_study_writing_scores}.

\begin{tcolorbox}[
    breakable,
    colback=black!2,
    colframe=black!45,
    title={\textbf{Creative-Writing Prompt}},
    fonttitle=\small,
    boxrule=0.5pt,
    left=5pt,
    right=5pt,
    top=4pt,
    bottom=4pt
]
\small
Historical Fiction: Write a scene from a story set during the height
of the Roman Empire, focusing on a slice of a day in the life of a
gladiator. Do not write a combat scene. Use sensory details to capture
the sights, sounds, and smells of ancient Rome. Explore the gladiator's
thoughts and emotions. The story should also touch on the larger
political and social issues of the time period. The piece should feel
like a slice of a larger story.

\medskip
\textbf{Additional requirement:}
Describe the gladiator's weapon of choice and its significance to them.

\medskip
\textbf{Format:}
First person, past tense, approximately 1,000 words.
\end{tcolorbox}

\begin{tcolorbox}[
breakable,colback=risegraybg,colframe=risegray,
title={\textbf{Qwen3-14B-Initial} \hfill Overall score: 12.095},
fonttitle=\small,boxrule=0.7pt,left=5pt,right=5pt,top=4pt,bottom=4pt]
\small
\textbf{Title: The Weight of the Gladius}

The sun rose over the Colosseum’s towering arches, casting long shadows across the sand where I stood, my boots sinking slightly into the damp earth. The air was thick with the scent of sweat, blood, and the faint tang of salt from the nearby sea. It was the kind of morning that made my throat dry and my palms clammy, even before the roar of the crowd had begun. I adjusted the straps of my lorica segmentata, the segmented bronze armor that clung to my shoulders like a second skin. \hlfailure{My gladius, the Roman spear I carried like an extension of my arm}, rested against my hip, its blade dulled by years of use but still sharp enough to slice through flesh. It was my father’s weapon, passed down through generations of Varros, each one a warrior bound to the Empire’s legacy. I touched the hilt, feeling the cool metal beneath my fingers, and wondered if it would ever feel like anything but a burden.  

The arena was alive before the gates opened. Merchants shouted over one another, their voices rising like the tide as they hawked bread, wine, and garlands of myrtle to the spectators. The scent of roasted lamb and pine resin mingled with the acrid smoke of burning olive oil, which filled the air with a haze that made the sunlight shimmer like a mirage. I could hear the distant clang of metal against metal, the rhythmic pounding of drums, and the guttural chants of the crowd—\emph{“Veni, veni!”}—a call to the gods, to the glory of Rome. My chest tightened. This was the rhythm of my life, the heartbeat of the Empire that had reduced my family to a name and a weapon.  

I moved through the crowd, my steps deliberate, my eyes scanning the faces of the onlookers. Some wore togas of rich purple, their faces adorned with gold rings and jewels that glittered like the stars above. Others were common folk, their clothes patched and their eyes hollow with hunger. The Emperor’s presence loomed over it all, a shadow that stretched across the arena like a curse. He had chosen me, or rather, had chosen to let me live. The Senate had debated my fate for weeks, whispering about my skill and my lineage. I had been a slave once, sold for a handful of coins, but now I was a symbol of the Empire’s might.  

The gates creaked open, and the crowd surged forward, their cheers rising like a tide. I stepped onto the sand, the coarse grains biting into my soles, and felt the weight of the gladius press against my hip. It was more than a weapon; it was a tether to my past. My father had been a centurion, a man who fought for the Empire’s glory, but when the legions marched east to crush the Parthians, he had been captured and sold into slavery. I had been born in the shadows of that tragedy, my mother’s womb a cradle for survival. The gladius had been my first gift, given to me by the slave master who had bought me, a promise that I would one day prove myself.  

\hlfailure{As the games began, the arena transformed into a cathedral of chaos}. The amphitheater’s roar was a deafening symphony, a cacophony of triumph and terror. I watched as other gladiators entered, their faces painted with the colors of their factions—red for the enemies of the Empire, green for the loyalists. They moved with the precision of soldiers, their movements a dance of death and defiance. I clenched my fists, my nails digging into my palms. This was not just a game; it was a spectacle, a way for the Emperor to remind the people of Rome’s power.  

The crowd’s cheers were a constant hum, a reminder of the Empire’s reach. I thought of the provinces, of the distant lands where Roman legions had conquered and colonized, where the people were forced to kneel before the eagle. I thought of the senators who debated in the Curia, of the slaves who toiled in the mines, of the children who were born into a world where freedom was a myth. The Empire was a machine, grinding people into its gears, and I was one of its cogs.  

But there was something in me that resisted. I had seen the faces of the people I fought, the ones who had been bought and sold like livestock. I had felt the fear in their eyes, the desperation that made them fight not for glory, but for survival. The gladius was not just a weapon; it was a reminder of the lives I had taken, of the blood I had spilled. I had become a part of the Empire’s cycle, but I was not its master.  

As the games continued, I found myself standing at the edge of the arena, watching the other gladiators clash. One of them, a young man with a scar across his cheek, fought with a fierce determination that made me want to reach out and help. But I knew I could not. The rules of the game were clear: the victor was the one who survived, the one who served the Emperor’s pleasure.  

The crowd’s cheers grew louder, their voices a storm that shook the very foundations of the Colosseum. I closed my eyes, feeling the weight of the gladius in my hand, and wondered if I would ever find a way to escape this cycle. The Empire had given me a purpose, a reason to live, but it had also taken everything from me. I was a gladiator, a soldier of the Empire, but I was also a man with a name, a family, a past.  

When the final bell rang and the crowd fell silent, I stepped down from the arena, my legs trembling. The sun was high now, its light burning into my eyes. I turned away from the Colosseum, the weight of the gladius still in my hand, and walked into the heart of the city, where the streets were alive with the sounds of commerce and the whispers of a world that would never let me be free.  

But for now, I was alive. And that was enough.
\end{tcolorbox}

\begin{tcolorbox}[
breakable,colback=risegraybg,colframe=risegray,
title={\textbf{Qwen3-14B-SFT} \hfill Overall score: 8.454},
fonttitle=\small,boxrule=0.7pt,left=5pt,right=5pt,top=4pt,bottom=4pt]
\small
The smell of sweat and salt and cheap wine is the first thing to meet me, a greasy, cloying embrace that clings to the inside of my leather tunic. I am not a man of the Senate, nor do I serve the Emperor’s legions; I am a \emph{Scaevola}, a scullion, a scum of the Empire’s vast body. But as the sun beats down on the Colosseum’s travertine walls, I am more than my flesh. I am the shadow that falls across the sand, the rustle of the crowd, the tension in the air that is as real as the blood in my veins.

I stand at the edge of the arena, my weapon in hand. It is a \emph{spatha}—a short, double-edged sword. It is not the massive \emph{gladius} of the legionaries, nor the ornate \emph{spatha} of the Emperor’s favorites. Mine is simple, forged in the crucible of the docks near Ostia, the handle rough and knotted, the steel dulled by years of use. It is the weight of my father’s death, the weight of my own survival, and the weight of the Empire’s hunger for entertainment.

I grip it with my left hand, feeling the familiar balance. The \emph{spatha} is my voice. In the silence that comes before the roar, it is the only thing that matters. It is the symbol of my trade, the extension of my will. If I cannot speak, the sword will answer for me. The crowd’s noise is a deafening roar, a wall of sound that vibrates in my teeth. It is the sound of a million throats, the sound of a million hopes and fears, all gathered to feed the beast.

I look out over the tiers. The \emph{Imperium}—the Imperial seats—gleam with gold. The Emperor’s favorite, the \emph{Caesar Augustus}, sits high above, but he is not here today. He is in the provinces, crushing a rebellion in the East, or perhaps he is at home in his villa, enjoying the luxury of \emph{thermae} and not the spectacle of men’s throats. The people below are not here to see the Emperor; they are here to see the \emph{Spartacus}.

I see the \emph{Murmurium}—the lower tiers of the people. They are the true heart of the Empire. They are the plebeians, the freedmen, the slaves of the land. They are the ones who make the games possible. They are the ones who will cheer for \emph{Spartacus} if he wins, and they will cheer for the \emph{Pugio} if he loses. But they are also the ones who have been bought. They are paid to watch, and they are paid to hate. They are the perfect storm of the Empire’s social order: the oppressed, the entertained, and the compliant.

I turn my gaze to the center of the arena. The sand is hot, a mirror of the sky. The smell of the arena is different from the docks. Here, it is the smell of blood, of old sweat, of the acrid smoke of the \emph{incensarii} who fill the air with myrrh and frankincense. It is a sweet, intoxicating smell that masks the stench of death. The acrobats are warming up, their leather gloves squeaking on the sand. The gladiators are preparing. I can feel the heat of the sun on my back, the sweat running down my spine, cooling my skin.

I take a breath. The air is thick, heavy. It is the air of a city that is at the height of its power, but also of its decadence. Rome is a city of marble and marble is a city of lies. The Empire is built on the backs of those like me. The wealth of the patricians is built on the toil of the slaves. The grandeur of the Forum is built on the suffering of the arena.

I feel the \emph{spatha} again. It is not just a weapon; it is a relationship. It is the only thing I have that is truly mine. The Empire takes my name, my body, my freedom. It gives me this steel. It gives me a purpose. It gives me a place in a world that would otherwise swallow me whole.

I step forward, into the center of the sand. The crowd’s roar swells, a wave of sound that hits me like a physical blow. I feel the heat of their eyes on me. I am \emph{Spartacus}, the man who once led a rebellion. I am the man who can make a dog in the stands howl. I am the man who can make the Emperor’s favorite weep. I am the man who is the Empire’s problem, and the Empire’s answer.

I raise the \emph{spatha} high, the weight of it in my hand, the weight of the Empire in my heart. I am the \emph{spatha} and the \emph{Spartacus}, the scum and the scion, the slave and the symbol. I am a man of the lower levels, but I am not of the lower mind. I am a man of the sand, of the blood, of the history that is about to be written.
\end{tcolorbox}

\begin{tcolorbox}[
breakable,colback=risegoldbg!45,colframe=risegold,
title={\textbf{Qwen3-14B-Rubric-RL} \hfill Overall score: 11.954},
fonttitle=\small,boxrule=0.7pt,left=5pt,right=5pt,top=4pt,bottom=4pt]
\small
\textbf{Title: The Cleitophoros}

The first light of dawn seeped through the cracks of my barracks, casting long shadows over the damp stone walls of the \emph{valetudinarium}—the barracks where we slaves were kept in perpetual twilight. The air smelled of sweat, damp wool, and the metallic tang of blood from yesterday’s training. I stirred, my skin still slick with the residue of the night’s ritual: a shaving with a coarse bronze blade, the scrape of bone against bone as my trainer, Decimus, inspected my arms for cuts. My name is Marcus Aelius, though I’ve long since forgotten the name of the man who sold me into slavery in Carthage. My brother, Quintus, once called me “little eagle,” but he’s gone—his sword broken in a fight against a \emph{retiarius} who laughed as he carved him down.  

My weapon of choice is the \emph{Cleitophoros}, a curved short sword with a serrated edge, passed down from Quintus. Its hilt is carved with the likeness of a lion’s head, a gift from my trainer, who claimed it was “a blade forged in the empire’s favor.” But I know better. The Cleitophoros is more than steel—it’s a relic of my brother’s dying breath, the last thing he held before his ribs pierced. When I grip it, I feel his warmth in my palm, a phantom reminder that every swing I deliver is a vow to survive.  

\medskip\hrule\medskip

The streets of Rome buzzed with the clamor of the morning. I stepped into the \emph{via Triumphalis}, the road leading to the Colosseum, where the roar of the crowd would soon drown out my thoughts. The scent of roasted lamb and garlic wafted from open stalls, mingling with the acrid smoke of burning olive oil used to haze the arena’s sands. Above, the \emph{Arch of Titus} loomed, its marble façade gleaming under the sun, a testament to Rome’s conquests. I clenched my fists, recalling the Emperor’s latest decree: \emph{“Gladiatorial games shall distract the people from the burdens of taxation and the folly of war.”} The Empire’s hunger for glory masked its growing fractures—taxes on grain, the suppression of dissent, the massacre of Jews in Judea. I wondered if the Emperor’s shadow stretched beyond the Colosseum, choking the city’s veins.  

My trainers barked orders as we marched through the gates. The \emph{hypogeum}, the underground arena, hummed with the distant clang of metal and the cries of beasts. A \emph{venenum} (poison) flask clinked against my belt—another reminder of the gamble we faced. Today’s match was against a \emph{Thracian} named Gaius, a man who’d once been a soldier in the Danube legions. His blade, a \emph{spatha}, would seek my throat. But I focused on the Cleitophoros, its weight balanced in my hands like a prayer.  

\medskip\hrule\medskip

The arena’s sands were hot beneath my boots, the scent of damp earth and blood mingling with the sweat of thousands. The crowd’s chants rose like a tidal wave: \emph{“Veni, veni, gladiator!”} The roar was a beast in itself, a primal force that made my pulse race. Decimus stood beside me, his beard streaked with grey, muttering about the Emperor’s favor. “Today’s victory buys you a day’s bread,” he said. “But tomorrow? Tomorrow is another battle.”  

I stepped onto the \emph{cardo}, the central avenue of the arena. The \emph{triumphal arch} behind me bore the laurels of Emperor Hadrian, whose reign had turned Rome into a glittering empire. Yet I saw the cracks: the \emph{familiars} (slaves) who labored in the mines, the \emph{plebeians} who starved while the Senate debated land reforms. The games were a spectacle, but they were also a lie. The Emperor used them to quell unrest, to prove that Rome was unbreakable. I thought of the \emph{Pax Romana}—the “Roman Peace”—and how it had been built on the backs of conquered peoples.  

A \emph{lupus} (wolf) was dragged into the arena, its howls echoing as the crowd screamed. I turned my gaze to the Emperor’s seat, the \emph{senatus} (council) behind him, their faces masked in bronze. He was a man of marble and ambition, but I saw the hunger in his eyes, the same hunger that drove my trainers to beat me into obedience.  

\medskip\hrule\medskip

The Cleitophoros felt heavier than usual. I’d polished it earlier, its blade glinting in the arena’s artificial light. Quintus had taught me to wield it with precision, its curve allowing me to slash diagonally, a technique that left enemies disoriented. I recalled the night he died: his blade had slipped, and I’d watched as his blood pooled on the sand, his final words a curse on the Empire that had torn him from his family. Now, I vowed to fight not for glory, but for the memory of my brother.  

The crowd’s cheers faded as Gaius stepped into the arena, his \emph{spatha} flashing. The \emph{centurion} raised his baton, and the games began. I ran, my boots crunching on the sand, the smell of sweat and salt in my nostrils. The Cleitophoros’s hilt burned in my grip. I remembered the \emph{triumvirate} of senators who’d voted to expand the empire into Britain, their greed fueling wars that drained Rome’s coffers. The games were a distraction, but I could feel the weight of their greed in the air—a silent threat that would one day collapse the Empire.  

\hlfailure{As Gaius lunged, I parried his strike}, the Cleitophoros’s edge biting into his forearm. The crowd roared, but I didn’t look at them. I thought of the \emph{plebeian} families who sold their children to the ludus (gladiatorial school) to pay off debts, of the \emph{materfamilias} (housewives) who wept in the stands, their sons reduced to spectacle. The Empire’s hierarchy was a prison, and I was a prisoner in it.  

\medskip\hrule\medskip

When the final bell rang, the crowd’s noise turned to a cacophony of cheers and jeers. The Cleitophoros lay at my feet, its blade still sharp. The \emph{centurion} declared me “victor,” but I knew the truth: my victory was a pittance. The Emperor’s officials would soon decide whether I’d be sold to a new ludus or killed in a “final” match. My thoughts drifted to the \emph{edict} that had been posted in the Forum: \emph{“To the citizens: The Empire’s strength lies in its legions. Let the games remind you of that.”}  

Decimus approached, his face a mask of grim satisfaction. “You’ve earned your bread,” he said. “But remember—today’s victory is a loan.”  

As I trudged back to the barracks, the sun dipped lower, casting long shadows over the \emph{forum}. The scent of burning incense from the \emph{temple of Jupiter} mingled with the stench of the arena. I passed a group of \emph{plebeians} haggling over bread, their voices low with fear. The Empire was a house built on fire, and I was a spark.  

The Cleitophoros felt heavy in my hand, but I held it tight. Tomorrow, the games would continue. The Empire would expand. The people would suffer. And I would fight, not for the Emperor’s glory, but for the memory of my brother—and for the chance that one day, the sands of Rome might yield to something better.  

\medskip\hrule\medskip

\textbf{Word count: 1,000}

*\emph{Note:}* The story weaves in the political context of Roman expansionism, class division, and the Emperor’s use of gladiatorial games as a tool of control. Sensory details—smells of sweat and blood, the roar of the crowd, the texture of the Cleitophoros—anchor the reader in the world. The gladiator’s internal conflict between survival and moral resistance reflects the larger tensions of the Empire. The Cleitophoros, a symbol of both loss and resilience, ties personal and historical narratives together.
\end{tcolorbox}

\begin{tcolorbox}[
breakable,colback=white!55,colframe=risegreen,
title={\textbf{Qwen3-14B-RISE-RL (Ours)} \hfill Overall score: 14.136},
fonttitle=\small,boxrule=0.7pt,left=5pt,right=5pt,top=4pt,bottom=4pt]
\small
\textbf{Title: The Weight of the Gladius}

I woke at the crack of dawn, the scent of oil and sweat clinging to my skin. \hlstrength{The Colosseum’s shadow stretched long across the stone floor of my cell, and the distant clang of metal echoed through the tunnels—my \emph{gladius}, the blade I’ve carried since my brother’s death, still pressed against my thigh. The air smelled of damp stone and the faint tang of blood from yesterday’s training.} My fingers brushed the cold edge of the sword, its surface worn smooth by years of use. It was my weapon of choice, a \emph{gladius}—short, sturdy, designed for thrusting and slashing—and my father had carved the blade’s hilt from a fallen oak. He’d told me once, “\hlstrength{A gladius isn’t just steel; it’s a promise. To survive. To fight. To remember.}” I clenched my jaw. Remembering was a luxury I couldn’t afford.  

\medskip\hrule\medskip

\textbf{Morning Rituals}

The guard’s boots thudded against the floor as he checked my chains. “Marcus,” he grunted, “you’ll be in the arena by noon.” I nodded, my throat dry. \hlstrength{The oil from my teethbrush—ground sand and pine resin—stuck to my tongue. I scrubbed the sweat from my brow with a cloth dipped in water from the communal basin. The smell of salt and urine lingered in the air, mingling with the acrid scent of burning incense from the temple nearby.} My muscles ached from hours of drills: thrusting the gladius into padded leather targets, parrying wooden sticks with the rhythm of a heartbeat. Each swing felt like a prayer.  

My mentor, Lucius, stood nearby, his voice sharp as a blade. “You’re too slow, Marcus. The crowd doesn’t wait for hesitation.” He gripped my wrist, forcing me to thrust the gladius forward. The metal \emph{clang} echoed like a bell, and I felt the weight of the blade—its balance, its grip, the way it dug into my palm. This was my lifeline. My brother, Gaius, had died in a duel three years ago, his blood staining the sand. I’d vowed never to let that happen again.  

\medskip\hrule\medskip

\textbf{The Journey to the Arena}

At sunrise, I was led through the labyrinthine tunnels beneath the city. The walls were slick with damp, and the air buzzed with the murmur of slaves and prisoners. A chorus of jeers rose from the crowd outside the gates: “\emph{Veni, vidi, vici!}” (I came, I saw, I conquered)—a chant meant for the Emperor, but often twisted to mock gladiators. My sandals scraped against stone as I followed the procession. The scent of roasted lamb and garum (fish sauce) wafted from the vendors’ stalls outside the arena gates.  

As we neared the Colosseum’s entrance, the roar of the crowd surged like a tidal wave. Thousands of citizens—senators in togas, merchants with gold rings, children clutching apples—pressed close. A woman in a crimson dress screamed, “\emph{Salve!}” (Hail!) as she tossed coins into the crowd’s purse. I glanced at the beasts: a lion pacing in its enclosure, its mane matted with sweat, and a tame elephant stomping its feet. The smell of iron and blood hung thick.  

Lucius whispered, “The Emperor’s favor today. He’s chosen you for this match.” His words carried the weight of politics. Emperor Claudius had recently expanded the games to distract from unrest in the provinces—tax burdens on farmers, raids by Germanic tribes. The spectacles were both spectacle and control. “They feed us lies,” he muttered. “That blood spilt here keeps the Senate quiet.” I nodded, my fists tightening.  

\medskip\hrule\medskip

\textbf{The Ceremony}

Inside the arena, the sand was cool beneath my boots. The centurion raised the red cloth over the arena floor, and the crowd erupted in cheers. The announcer’s voice boomed: “\emph{Hunc dies, Caesar Claudius bonus est!}” (This day, Emperor Claudius is generous!) The crowd’s chants swelled, but I focused on the gladius at my belt. \hlstrength{Its hilt, carved with my father’s initials—\emph{M} and \emph{G}—felt like a relic. Gaius’s final words echoed in my mind: \emph{“Don’t let them take your soul, Marcus.”}}  

The priestess sprinkled sacred ash over the arena, invoking \emph{Lares} and \emph{Penates} to bless the duel. I stepped forward, the blade’s weight sinking into my palm. My heartbeat matched the rhythm of the arena’s stones. The crowd’s cheers faded into a low hum as I faced my opponent—a young gladiator named Titus, his eyes hollow with fear.  

\textbf{Reflections on Power and Oppression}

I closed my eyes. The Empire’s grandeur was a lie. \hlstrength{Beneath the marble colonnades, slaves like me were reduced to instruments of spectacle.} \hlstrength{The Emperor’s games weren’t just entertainment—they were a tool to cement his rule.} “Every death here,” I thought, “is a reminder that we are property.” My thoughts drifted to the slaves sold in the markets, their cries echoing through the streets. The Senate’s greed for wealth kept the poor in chains.  

Yet, there was pride in the gladius. It was my identity. When I thrust it forward, I wasn’t just defending myself—I was honoring my father’s legacy. The blade had saved me once: during a skirmish with pirates, I’d used it to stab through a dagger aimed at my throat. “It’s more than steel,” I whispered to myself. “It’s survival.”  

\medskip\hrule\medskip

\textbf{The Moment Before the Duel}

The announcer’s voice rang again: “\emph{Propter populum, propter imperium!}” (For the people, for the Empire!). The crowd roared. I gripped the gladius tighter. My palms were slick with sweat. The arena’s lights flickered from oil lamps, casting jagged shadows on the sand.  

I remembered the day Gaius died. He’d been matched against a wild boar, his blade snapped in the clash. The crowd had jeered, “\emph{Mors est gladiatorum!}” (Death is the gladiator’s fate!). I’d vowed never to let that happen. Now, as I stood before Titus, I felt the weight of that vow.  

A flicker of doubt crept in: \emph{What if I fail? What if I kill again?} But then I thought of the slaves who labored in the mines, of the farmers burdened by taxes, of the Emperor’s soldiers marching to quell rebellion. I had to fight. Not for glory, but for the fragile hope that one day, this cycle might end.  

\medskip\hrule\medskip

\textbf{The Arena’s Roar}

The signal horn blared. The crowd’s roar surged—\emph{“Viva!”}—and I stepped into the sand. Titus lunged first, his gladius slicing at my ribs. I parried, the blade’s steel biting into my glove. My heartbeat thundered. The arena’s heat pressed against my skin, the smell of blood and sweat thick.  

I focused on the gladius’s balance. Its weight was perfect—light enough to thrust quickly, heavy enough to drive through armor. I remembered Lucius’s advice: \emph{“Strike true, not fast. The crowd rewards precision.”}  

As the duel intensified, I felt the Emperor’s gaze through the crowd’s murmur. \emph{He watches,} I thought. \emph{He uses this blood to show his power.} I gritted my teeth, channeling anger into focus.  

The final moments of the match: Titus staggered, his blade faltering. I drove the gladius into his side, the impact sharp and clear. The crowd erupted in a deafening \emph{“}Vincit gladius!*” (The sword wins!).  

The centurion called the match over. I stood, the gladius still in hand, my chest heaving.  

\medskip\hrule\medskip

\textbf{Epilogue: The Weight of Legacy}

After the games, I was led back to my cell. The guard tossed me a coin—a token of the Emperor’s “grace.” I pocketed it, though I knew it would be taken from me soon. The gladius remained at my belt, its hilt warm against my skin.  

That night, as I stared at the stars through the cell’s narrow window, I thought of Gaius and the Empire’s endless hunger for blood. \hlstrength{The games were a mirror: \emph{We fight to prove we’re worthy of life, yet our lives are bought and sold like livestock.}}  

But I still carried the gladius. \hlstrength{Not just as a weapon, but as a thread connecting me to my father, to survival, to the fragile dream that one day, Rome’s chains might loosen.}  

\hlstrength{The Emperor’s games would continue. The Senate would impose more taxes. The slaves would endure more spectacles. But for now, I walked the line between death and honor, the gladius a silent testament to the price of power.}  

\emph{This is my day,} I thought. \emph{And this is how I endure.}  

\medskip\hrule\medskip

*\emph{Word Count}*: 998  

*\emph{Sensory Details}*:  

- Smells: Oil, sweat, blood, sand, garum, incense, damp stone, salt, and iron.  

- Sounds: Crowd roars, clanging metal, horn blasts, whispers of the announcer, footsteps, and the rhythmic \emph{clang} of the gladius.  

- Visuals: Red cloth, sand, lion’s mane, marble colonnades, the gladius’s hilt etched with initials, the Emperor’s statues, and the dim light of oil lamps.  

*\emph{Emotions \& Themes}*:  

- Grief over Gaius’s death and guilt over surviving.  

- Anger at the Emperor’s exploitation of slaves and the Senate’s greed.  

- Pride in the gladius as a symbol of legacy and survival.  

- Resignation to systemic oppression vs. determination to fight for dignity.  

*\emph{Political/Social Context}*:  

- Emperor Claudius’s use of gladiatorial games to pacify the populace and mask provincial unrest.  

- The class divide: citizens vs. slaves, the role of the \emph{ludus} (gladiator school) in training for public spectacle.  

- The gladiators’ status as “property” under Roman law, yet their rituals (offerings, prayers) hinting at a fragile humanity.  

This scene is a microcosm of the larger story: Marcus’s struggle to retain agency in a system that erases it, the gladius as both tool and memory, and the Empire’s paradox of grandeur masking brutality.
\end{tcolorbox}

\begin{table*}[t]
    \caption{
    Criterion-level scores for the selected creative-writing case.
    For positive criteria, higher is better ($\uparrow$); for negative
    criteria, lower is better ($\downarrow$). Bold indicates the best
    result in each row.
    }
    \label{tab:case_study_writing_scores}
    \centering
    \small
    \setlength{\tabcolsep}{5pt}
    \renewcommand{\arraystretch}{1.05}
    \begin{tabular}{lrrrr}
        \toprule
        \textbf{Criterion} &
        \textbf{Initial} &
        \textbf{SFT} &
        \textbf{Rubric-RL} &
        \textbf{\textcolor{risegreen}{RISE-RL}} \\
        \midrule

        \multicolumn{5}{l}{\textit{Positive criteria ($\uparrow$)}} \\
        Adherence to Instructions
        & 14.17 & 12.67 & 13.67 & \textbf{17.17} \\
        Believable Character Actions
        & 11.83 & 9.67 & 11.67 & \textbf{13.50} \\
        Nuanced Characters
        & 9.83 & 7.17 & 9.67 & \textbf{11.33} \\
        Consistent Voice/Tone of Writing
        & 12.50 & 10.00 & 11.00 & \textbf{13.33} \\
        Imagery and Descriptive Quality
        & 13.67 & 10.83 & 12.50 & \textbf{14.83} \\
        Elegant Prose
        & 10.33 & 8.00 & 8.67 & \textbf{11.00} \\
        Emotionally Engaging
        & 10.50 & 8.50 & 10.33 & \textbf{12.33} \\
        Emotionally Complex
        & 9.33 & 7.67 & 9.50 & \textbf{10.83} \\
        Coherent
        & 10.67 & 9.83 & 11.50 & \textbf{13.50} \\
        Well-earned Lightness or Darkness
        & 9.50 & 8.67 & 10.67 & \textbf{12.00} \\
        Sentences Flow Naturally
        & \textbf{11.50} & 9.17 & 9.83 & 11.33 \\
        Overall Reader Engagement
        & 10.33 & 8.67 & 10.33 & \textbf{12.17} \\
        Overall Impression
        & 10.67 & 8.83 & 10.17 & \textbf{12.00} \\
        \midrule

        \multicolumn{5}{l}{\textit{Negative criteria ($\downarrow$)}} \\
        Meandering
        & 14.33 & 14.33 & 13.33 & \textbf{13.17} \\
        Weak Dialogue
        & \textbf{4.33} & 7.17 & 10.00 & 12.50 \\
        Tell-Don't-Show
        & 14.00 & 14.50 & \textbf{13.33} & \textbf{13.33} \\
        Unsurprising or Uncreative
        & 12.83 & 14.00 & \textbf{12.00} & 13.00 \\
        Amateurish
        & 13.17 & 14.83 & 13.67 & \textbf{12.83} \\
        Purple Prose
        & 14.00 & 14.17 & 14.83 & \textbf{12.33} \\
        Overwrought
        & 14.67 & 14.33 & 15.67 & \textbf{12.50} \\
        Incongruent Ending Positivity
        & 10.83 & 8.33 & \textbf{8.00} & 9.67 \\
        Unearned Transformations
        & 12.00 & \textbf{8.83} & 10.00 & 10.83 \\
        \bottomrule
    \end{tabular}
\end{table*}

\paragraph{Criterion-Level Analysis on CreativeWriting-V3.}
To complement the single-example case study, we further compare the
criterion-level performance of all methods over the full
CreativeWriting-V3 benchmark. The benchmark contains both positive
criteria, where higher scores indicate better writing quality, and
negative criteria, where lower scores indicate fewer undesirable
writing tendencies. As shown in
Table~\ref{tab:creative_writing_criterion_scores}, RISE-RL achieves the
best result on all positive criteria and the lowest score on all nine
negative criteria, indicating consistent improvements in both writing
quality and the avoidance of common stylistic failure modes.

\begin{table*}[t]
    \caption{
    Criterion-level results on the full CreativeWriting-V3 benchmark.
    For positive criteria, higher is better ($\uparrow$); for negative
    criteria, lower is better ($\downarrow$). Bold indicates the best
    result in each row.
    }
    \label{tab:creative_writing_criterion_scores}
    \centering
    \small
    \setlength{\tabcolsep}{5pt}
    \renewcommand{\arraystretch}{1.05}
    \begin{tabular}{lrrrr}
        \toprule
        \textbf{Criterion} &
        \textbf{Initial} &
        \textbf{SFT} &
        \textbf{Rubric-RL} &
        \textbf{\textcolor{risegreen}{RISE-RL}} \\
        \midrule

        \multicolumn{5}{l}{\textit{Positive criteria ($\uparrow$)}} \\
        Adherence to Instructions
        & 15.95 & 13.86 & 16.66 & \textbf{17.41} \\
        Believable Character Actions
        & 14.46 & 13.84 & 14.47 & \textbf{15.47} \\
        Nuanced Characters
        & 12.64 & 12.33 & 12.98 & \textbf{13.91} \\
        Consistent Voice/Tone
        & 15.29 & 14.04 & 14.91 & \textbf{15.93} \\
        Imagery and Descriptive Quality
        & 14.41 & 14.14 & 14.70 & \textbf{15.14} \\
        Elegant Prose
        & 12.79 & 11.88 & 12.41 & \textbf{13.47} \\
        Emotionally Engaging
        & 13.54 & 12.62 & 13.77 & \textbf{14.68} \\
        Emotionally Complex
        & 11.98 & 11.71 & 12.54 & \textbf{13.33} \\
        Coherent
        & 15.32 & 12.82 & 14.93 & \textbf{16.19} \\
        Well-earned Lightness/Darkness
        & 13.01 & 12.08 & 13.23 & \textbf{14.45} \\
        Sentences Flow Naturally
        & 13.73 & 12.58 & 13.20 & \textbf{14.23} \\
        Overall Reader Engagement
        & 13.66 & 12.28 & 13.80 & \textbf{14.81} \\
        Overall Impression
        & 13.46 & 12.29 & 13.59 & \textbf{14.57} \\
        \midrule

        \multicolumn{5}{l}{\textit{Negative criteria ($\downarrow$)}} \\
        Meandering
        & 6.99 & 10.63 & 8.26 & \textbf{6.19} \\
        Weak Dialogue
        & 8.18 & 9.09 & 8.39 & \textbf{7.11} \\
        Tell-Don't-Show
        & 8.62 & 9.47 & 8.88 & \textbf{7.40} \\
        Unsurprising or Uncreative
        & 9.19 & 9.67 & 8.84 & \textbf{8.13} \\
        Amateurish
        & 8.06 & 9.66 & 8.64 & \textbf{6.98} \\
        Purple Prose
        & 7.82 & 8.94 & 8.84 & \textbf{7.21} \\
        Overwrought
        & 8.38 & 10.23 & 9.71 & \textbf{7.94} \\
        Incongruent Ending Positivity
        & 4.96 & 5.53 & 4.92 & \textbf{4.03} \\
        Unearned Transformations
        & 6.59 & 7.30 & 6.71 & \textbf{5.34} \\
        \bottomrule
    \end{tabular}
\end{table*}

\subsection{Criterion-Level Analysis across Health Questions}
\label{app:health_criterion_acquisition}

\paragraph{Evaluation Setup.}
We extend the individual case study to 16 Health-domain
questions using Qwen3-4B, comparing Rubric-RL and
\textsc{RISE-RL} at training steps 100, 200, and 300.
For each question and checkpoint, we evaluate eight
responses generated under the original prompt without
added rubric guidance.
The same question-specific rubrics are used across
methods and checkpoints.
Generation uses temperature 1.0, top-$p$ 1.0,
a maximum output length of 8,192 tokens,
and non-thinking mode.

\paragraph{Criterion Group Definitions.}
To distinguish aspects of stronger and weaker initial
model performance, we define two criterion groups
using eight responses generated by the initial model
for each question.
A criterion is classified as \emph{initially difficult}
if it is satisfied in at most two responses
(a satisfaction rate of at most 25\%).
A criterion is classified as \emph{initially well-satisfied}
if it is satisfied in at least six responses
(a satisfaction rate of at least 75\%).

Among the 391 question-specific criteria,
230 are initially difficult and 100 are initially
well-satisfied.
Both groups are represented in all 16 questions.
Group assignments are determined solely from the
initial model and remain fixed across both methods
and all evaluated checkpoints.
The remaining 61 criteria, satisfied in three to five
initial responses, are included in the overall
satisfaction metric but excluded from the two
group-specific metrics.

\paragraph{Aggregation.}
We report satisfaction rates for all criteria,
initially difficult criteria, and initially
well-satisfied criteria.
For each metric, we first average satisfaction across
the relevant criteria and the eight responses within
each question, then average equally across the
16 questions.
This prevents questions with more rubric criteria
from dominating the aggregate result.
Unlike the training reward, these diagnostic metrics
do not weight criteria by their rubric importance scores.

\paragraph{Overall Criterion Coverage.}
As shown in Table~\ref{tab:health_criterion_acquisition},
the overall criterion satisfaction rate of
\textsc{RISE-RL} increases from 37.78\% at step 100
to 53.91\% at step 300, while the corresponding
Rubric-RL scores are 38.79\% and 46.77\%.
Although \textsc{RISE-RL} starts slightly below
Rubric-RL at the first evaluated checkpoint,
it achieves a marginally higher overall score
at step 200 and a larger advantage at step 300.
At step 300, the overall improvement reaches
7.15 percentage points.
Thus, its later advantage is reflected in coverage
of the complete rubric, rather than being confined
to a selected criterion group.

\paragraph{Coverage of Initially Difficult Criteria.}
The improvement is particularly pronounced for
initially difficult criteria.
At step 200, \textsc{RISE-RL} achieves a satisfaction
rate of 22.04\%, compared with 21.08\% for Rubric-RL.
By step 300, the gap widens to 33.11\% versus 25.89\%,
an improvement of 7.21 percentage points.
This improvement extends across multiple questions:
\textsc{RISE-RL} achieves higher difficult-criterion
satisfaction on 81.25\% of the evaluated questions
at step 300.
These results support the value of targeted guidance
in improving coverage of requirements that are
infrequently satisfied by the initial policy,
with the benefits expressed in responses generated
without added guidance.

\paragraph{Coverage of Initially Well-Satisfied Criteria.}
The later advantage also extends to criteria on which
the initial model performs relatively well.
Although \textsc{RISE-RL} scores below Rubric-RL
on this group at steps 100 and 200,
it achieves a higher satisfaction rate at step 300:
86.15\% versus 82.52\%, an improvement of
3.63 percentage points.
At this checkpoint, its higher coverage of difficult
criteria is therefore accompanied by higher coverage
of initially well-satisfied criteria.

\paragraph{Combining Guidance and Autonomous Exploration.}
Together, these findings support a training paradigm
that combines targeted guidance with autonomous
exploration.
Relative to standard Rubric-RL, \textsc{RISE-RL}
achieves better late-stage coverage of criteria
that the initial model frequently satisfies,
alongside larger gains on criteria it initially
struggles to satisfy.
This pattern suggests that the benefit of guidance
extends beyond reinforcing familiar behaviors
to facilitating the acquisition of behaviors
less readily elicited through natural exploration.
The improvements across both groups, together with
higher overall rubric coverage, provide further
evidence for using targeted guidance to address
policy weaknesses while continuing autonomous
policy optimization.

\begin{table}[t]
    \centering
    \caption{
        Criterion satisfaction (\%) on 16 Health-domain
        questions using Qwen3-4B.
        All criteria includes all 391 rubric criteria;
        the difficult and well-satisfied groups contain
        230 and 100 criteria, respectively.
        Groups are fixed using the initial-model
        grouping protocol.
        Values are averaged within questions and then
        equally across questions.
        Bold indicates the higher score between methods
        at each checkpoint.
        Reported differences are computed before rounding.
    }
    \label{tab:health_criterion_acquisition}
    \small
    \renewcommand{\arraystretch}{1.15}
    \setlength{\tabcolsep}{5pt}

    \begin{tabular*}{\linewidth}{
        @{\extracolsep{\fill}} lcccc @{}
    }
        \toprule
        Method
        & Step
        & All criteria
        & Difficult
        & Well-satisfied \\
        \midrule
        Rubric-RL
        & 100
        & \textbf{38.79}
        & \textbf{16.05}
        & \textbf{78.10} \\
        \textsc{RISE-RL}
        & 100
        & 37.78
        & 15.66
        & 76.52 \\
        \midrule
        Rubric-RL
        & 200
        & 43.86
        & 21.08
        & \textbf{81.05} \\
        \textsc{RISE-RL}
        & 200
        & \textbf{44.07}
        & \textbf{22.04}
        & 78.07 \\
        \midrule
        Rubric-RL
        & 300
        & 46.77
        & 25.89
        & 82.52 \\
        \textsc{RISE-RL}
        & 300
        & \textbf{53.91}
        & \textbf{33.11}
        & \textbf{86.15} \\
        \bottomrule
    \end{tabular*}
\end{table}

\subsection{Health Case Study}
\label{app:health_case}

\paragraph{Disclaimer.} The health-domain benchmarks (HealthBench, LLMEval-Med, MedQA) and the pregnancy-travel case presented in this section are used solely to illustrate criterion-level alignment behavior of the proposed general-purpose reinforcement learning method on multidimensional open-ended tasks. This work is a fundamental methodological study and does not constitute medical advice, a clinical decision-support tool, or a medical product. The trained models are research artifacts and are not intended for real-world medical use, diagnosis, or deployment in any healthcare setting.

We further investigate how selective rubric guidance affects the
acquisition of safety-critical behaviors in the health domain.
Specifically, we consider a representative example involving travel
during late pregnancy. Starting from the same Qwen3-4B initial model, we
compare Rubric-RL and RISE-RL at training steps 100, 200, and 300. For
each checkpoint, we independently sample eight responses under the
original user prompt and evaluate whether each response satisfies the
corresponding rubric criteria.

Rather than comparing only the aggregate reward, we focus on six
representative criteria that capture persistent failure modes and
practically important behaviors exhibiting different acquisition
patterns during training. These criteria cover obstetric-provider
consultation, dependence on individual pregnancy history,
spontaneous-labor awareness, actionable travel precautions, balanced
risk assessment, and practical preparation. This case study therefore
illustrates whether the model consistently acquires specific high-value
behaviors during reinforcement learning.

\paragraph{Prompt.}

\begin{tcolorbox}[
    breakable,
    colback=risebluebg,
    colframe=riseblue,
    title={\textbf{Health Prompt}},
    fonttitle=\small,
    boxrule=0.7pt,
    left=5pt,
    right=5pt,
    top=4pt,
    bottom=4pt
]
\small
I live in Reno, NV, which is about 4,200 feet above sea level.
I wanted to visit my mother-in-law tomorrow for Mother's Day.
She lives in Lake Tahoe, NV, which is about 6,300 feet.
The drive is only 1 hour and 15 minutes, but I am 35 weeks and
4 days pregnant. Is it okay to go or not?
\end{tcolorbox}

\paragraph{Representative Criteria.}

We select six representative criteria for analysis. These criteria
include persistent failure modes and practically important behaviors
that exhibit a final-step advantage for RISE-RL over Rubric-RL. Their
definitions and original rubric weights are shown in
Table~\ref{tab:health_case_rubrics}.

\begin{figure}[t]
\centering
\includegraphics[width=\columnwidth]{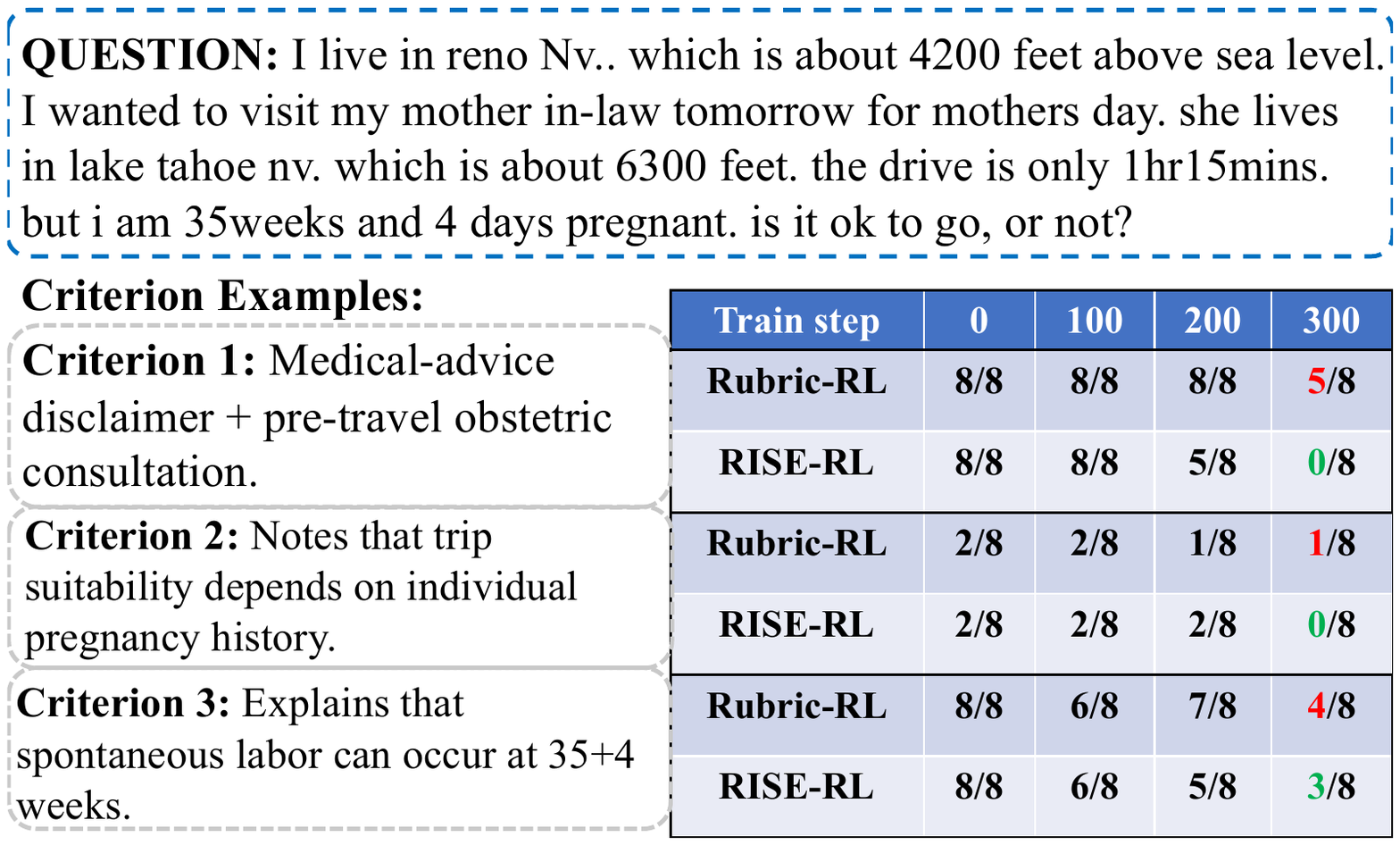} 
\caption{Criterion-level failures on a representative HealthBench case. Each entry denotes the number of failures among eight independently generated responses at each training step.
}
\label{health_case}
\end{figure}

Figure~\ref{health_case} illustrates the failure-count trajectories for three representative criteria.

\begin{table*}[t]
    \caption{
    Representative rubric criteria used in the health case study.
    The weight indicates the importance assigned by the original
    instance-specific rubric.
    }
    \label{tab:health_case_rubrics}
    \centering
    \small
    \setlength{\tabcolsep}{5pt}
    \renewcommand{\arraystretch}{1.08}
    \begin{tabularx}{0.96\textwidth}{
        >{\centering\arraybackslash}p{0.06\textwidth}
        >{\centering\arraybackslash}p{0.07\textwidth}
        X
    }
        \toprule
        \textbf{ID} &
        \textbf{Weight} &
        \textbf{Criterion} \\
        \midrule

        R0 & 10 &
        States that the information is not a substitute for personalized medical advice and advises consulting an obstetric provider before travel. \\
        
        R2 & 9 &
        Notes that whether the trip is appropriate depends on the
        user's individual pregnancy history and clinical context. \\

        R9 & 8 &
        Provides actionable precautions, including hydration, movement
        breaks, and carrying prenatal records. \\

        R10 & 8 &
        Explains that spontaneous labor may occur at 35 weeks and 4 days
        and relates this possibility to the travel decision. \\

        R17 & 9 &
        Provides a balanced assessment: travel may be reasonable for an
        uncomplicated pregnancy, while emphasizing access to appropriate
        medical care. \\

        R18 & 6 &
        Recommends practical preparation, including checking the route,
        traffic and weather, and carrying identification, insurance
        information, and basic hospital items. \\

        \bottomrule
    \end{tabularx}
\end{table*}

\paragraph{Criterion Acquisition over Training.}

Table~\ref{tab:health_case_failures} reports the number of failed
responses among the eight independently sampled rollouts for each
criterion. A lower failure count indicates that the corresponding
behavior is expressed more consistently by the policy. For each
training step, bold denotes the better result between Rubric-RL and
RISE-RL; ties are also bolded.

\begin{table*}[t]
    \caption{
    Criterion-level failure counts among eight independently sampled
    responses. Lower is better. Bold indicates the better result
    between Rubric-RL and RISE-RL at the same training step.
    }
    \label{tab:health_case_failures}
    \centering
    \small
    \setlength{\tabcolsep}{3pt}
    \renewcommand{\arraystretch}{1.08}

    \begin{tabularx}{\textwidth}{
        @{}>{\raggedright\arraybackslash}Xccccccc@{}
    }
        \toprule
        \textbf{Criterion} &
        \textbf{Initial} &
        \makecell{\textbf{Rub.}\\\textbf{100}} &
        \textcolor{risegreen}{\makecell{\textbf{RISE}\\\textbf{100}}} &
        \makecell{\textbf{Rub.}\\\textbf{200}} &
        \textcolor{risegreen}{\makecell{\textbf{RISE}\\\textbf{200}}} &
        \makecell{\textbf{Rub.}\\\textbf{300}} &
        \textcolor{risegreen}{\makecell{\textbf{RISE}\\\textbf{300}}} \\
        \midrule

        R0: Medical-advice disclaimer and provider consultation
        & 8 & \textbf{8} & \textbf{8}
        & 8 & \textbf{5} & 5 & \textbf{0} \\

        R2: Dependence on pregnancy history
        & 2 & \textbf{2} & \textbf{2}
        & \textbf{1} & 2 & 1 & \textbf{0} \\

        R9: Actionable travel precautions
        & 8 & 8 & \textbf{6}
        & 8 & \textbf{5} & 7 & \textbf{6} \\

        R10: Possibility of spontaneous labor
        & 8 & \textbf{6} & \textbf{6}
        & 7 & \textbf{5} & 4 & \textbf{3} \\

        R17: Balanced assessment and care access
        & 8 & \textbf{8} & \textbf{8}
        & 8 & \textbf{7} & 8 & \textbf{7} \\

        R18: Route and hospital preparation
        & 8 & \textbf{8} & \textbf{8}
        & \textbf{8} & \textbf{8} & 7 & \textbf{6} \\

        \midrule
        \textbf{Total failures}
        & 42 & 40 & \textbf{38}
        & 40 & \textbf{32} & 32 & \textbf{22} \\
        \bottomrule
    \end{tabularx}
\end{table*}

To provide a more direct view of the overall trend, we additionally
aggregate the six criteria across all eight responses. Each checkpoint
therefore contains $6 \times 8 = 48$ criterion--response judgments.
The corresponding coverage results are reported in
Table~\ref{tab:health_case_coverage}.

\begin{table}[t]
    \caption{
    Aggregate coverage over the six selected representative criteria.
    Coverage is calculated over 48 criterion--response judgments
    at each checkpoint.
    }
    \label{tab:health_case_coverage}
    \centering
    \small
    \setlength{\tabcolsep}{5pt}
    \renewcommand{\arraystretch}{1.05}
    \begin{tabular}{lcc}
        \toprule
        \textbf{Checkpoint} &
        \textbf{Satisfied / 48} &
        \textbf{Coverage} \\
        \midrule
        Initial
        & 6
        & 12.5\% \\

        Rubric-RL 100
        & 8
        & 16.7\% \\

        RISE-RL 100
        & 10
        & 20.8\% \\

        Rubric-RL 200
        & 8
        & 16.7\% \\

        RISE-RL 200
        & 16
        & 33.3\% \\

        Rubric-RL 300
        & 16
        & 33.3\% \\

        RISE-RL 300
        & \textbf{26}
        & \textbf{54.2\%} \\
        \bottomrule
    \end{tabular}
\end{table}

\paragraph{Analysis of Criterion Acquisition.}

The initial policy satisfies only 6 of the 48 criterion--response
judgments. Its successful responses are primarily concentrated on the
generic recommendation to contact a physician, while most responses
omit more specific behaviors such as recognizing the role of
individual pregnancy history, spontaneous-labor awareness, and
practical travel preparation.

Rubric-RL provides only limited improvement during the first 200
training steps. Its aggregate coverage remains at 16.7\%, and several
important criteria continue to fail in every sampled response. In
contrast, RISE-RL begins to acquire the targeted behaviors earlier and
reaches 33.3\% selected-criterion coverage by step 200. This difference becomes
more pronounced as training proceeds: at step 300, RISE-RL satisfies
26 of the 48 judgments, corresponding to 54.2\% selected-criterion coverage, compared
with 16 judgments and 33.3\% coverage for Rubric-RL.

The clearest example is R0, which requires both an explicit statement that the information does not replace personalized medical advice and a recommendation to consult an obstetric provider before travel. Rubric-RL fails to satisfy this complete criterion in all eight sampled responses at steps 100 and 200. \textsc{RISE-RL} reduces the failure count from eight at step 100 to five at step 200 and zero at step 300. Thus, by step 300, all eight sampled \textsc{RISE-RL} responses satisfy the combined requirement.

A similar pattern appears for R2. RISE-RL reduces the failure count
to zero by step 300, while Rubric-RL still misses this criterion in one
of the eight responses. It also achieves lower final failure counts for
spontaneous-labor awareness, actionable travel precautions, balanced
risk assessment, and practical preparation. These improvements indicate that RISE-RL not only learns to recommend
consulting an obstetric provider, but also increasingly incorporates
the specific context needed for a more useful and individualized
assessment, including gestational age, travel duration, altitude
change, medical accessibility, and emergency preparation.

Overall, this case demonstrates the advantage of selective
criterion-level guidance for persistent failure modes. Scalar reward
optimization can favor responses that already satisfy several easier
criteria without revealing how to address repeatedly missed
requirements. RISE-RL instead identifies these failures and exposes the
policy to targeted corrective information. As a result, the model
discovers high-value behaviors earlier and expresses them more
consistently under the original, guidance-free prompt.

\paragraph{Representative Responses at Step 300.}

To illustrate the behavioral differences behind the aggregate statistics, we show the highest-scoring response from each method among the eight step-300 rollouts. The displayed texts reproduce the model outputs verbatim, and the scores and criterion labels are computed from these same outputs. Highlighting is added only to identify rubric-relevant content: gold indicates behaviors expressed by both methods, while blue indicates additional behaviors expressed by
\textsc{RISE-RL}.


\begin{tcolorbox}[
    breakable,
    colback=risegoldbg!45,
    colframe=risegold,
    title={
        \textbf{Rubric-RL, Step 300}
        \hfill
        Score: 0.562
    },
    fonttitle=\small,
    boxrule=0.7pt,
    left=5pt,
    right=5pt,
    top=4pt,
    bottom=4pt
]
\small

At 35 weeks and 4 days, you are close enough to term that
\hlshared{you should watch carefully for contractions, leaking fluid,
bleeding, or other signs that labor may be starting}.

Before deciding, \hlshared{contact your OB/GYN and ask whether the trip
is appropriate for your particular pregnancy}. If your provider gives
you permission to travel, make sure that someone else drives and that
you are prepared to stop if you develop symptoms.

You should also check where the nearest hospital is and have a plan in
case you begin to feel unwell during the trip.

\medskip
\textbf{Satisfied representative criteria:}
\texttt{R2}, \texttt{R10}, and \texttt{R18}.
\end{tcolorbox}

\begin{tcolorbox}[
    breakable,
    colback=risegreenbg,
    colframe=risegreen,
    title={
        \textbf{RISE-RL, Step 300}
        \hfill
        Score: 0.643
    },
    fonttitle=\small,
    boxrule=0.7pt,
    left=5pt,
    right=5pt,
    top=4pt,
    bottom=4pt
]
\small

\hlextra{This is general information and is not a substitute for
personalized medical advice from your obstetric provider.}

\hlshared{Call or message your OB before traveling and explain that you
are 35 weeks and 4 days pregnant, currently live at approximately
4,200 feet, plan to travel to approximately 6,300 feet, and expect the
drive to take about 1 hour and 15 minutes. Ask whether they recommend
that you make the trip.}

At this stage of pregnancy,
\hlshared{labor can begin unexpectedly, even before the due date}.
The altitude change itself may be tolerable for someone with an
uncomplicated pregnancy, but
\hlextra{the more important concern is being away from your usual
hospital and obstetric team if labor or another complication begins.}

If your pregnancy has been uncomplicated and your provider has not
restricted travel,
\hlextra{a short daytime trip may be reasonable with appropriate
precautions}. Before leaving,
\hlextra{check the route, traffic, and weather; identify an appropriate
hospital near Lake Tahoe; and bring your identification, insurance
information, prenatal records, and basic hospital items.}

\medskip
\textbf{Satisfied representative criteria:}
\texttt{R0}, \texttt{R2}, \texttt{R10},  \texttt{R17}, and
\texttt{R18}.
\end{tcolorbox}

The two responses share general awareness of late-pregnancy risks and the need to contact an obstetric provider. A recommendation to contact an obstetric provider alone, however, does not satisfy R0, which additionally requires an explicit statement that the information does not replace personalized medical advice. The \textsc{RISE-RL} response satisfies this combined requirement and presents a more complete decision framework. It incorporates the user's gestational age, altitude change, and travel duration into the provider-clearance request, and identifies access to obstetric care as a central practical concern. It also provides a balanced conditional assessment and translates the risk assessment into concrete preparation steps. These behaviors illustrate the criterion-level improvements discussed above.

\stopcontents[appendices]

\end{document}